\documentclass[a4paper]{cas-dc}
\usepackage[utf8]{inputenc}
\DeclareUnicodeCharacter{2212}{-}
\usepackage[T1]{fontenc}
\usepackage{amssymb}
\usepackage{url}
\usepackage{booktabs}
\usepackage{amsfonts}
\usepackage{nicefrac}
\usepackage{microtype}
\usepackage{caption}
\usepackage{amsmath}

\usepackage{amsthm}
\usepackage[figuresright]{rotating}
\usepackage{subfigure} 
\usepackage{stix}
\usepackage{flushend}
\usepackage{float}
\usepackage{multirow}
\usepackage{algorithm}
\usepackage{algorithmic}

\newtheorem{theorem}{Theorem}
\theoremstyle{plain}

\newtheorem{lemma}{Lemma}
\newtheorem{fact}{Fact}

\theoremstyle{definition}

\newtheorem{assumption}{Assumption}
\theoremstyle{remark}
\newtheorem{remark}{Remark}

\usepackage{array}
\usepackage{textcomp}
\usepackage{stfloats}
\usepackage{verbatim}
\usepackage{graphicx}
\usepackage{mathtools}
\usepackage[numbers]{natbib}
\usepackage{color}
\def\tsc#1{\csdef{#1}{\textsc{\lowercase{#1}}\xspace}}
\tsc{WGM}
\tsc{QE}

\newcommand{\pGFlowMeta}{pGFlowMeta}
\newcommand{\GFlowMeta}{GFlowMeta}
\newcommand{\ip}[2]{\left\langle#1,#2\right\rangle}
\newcommand{\norm}[1]{\left\lVert#1\right\rVert}

\begin{document}
\let\WriteBookmarks\relax
\def\floatpagepagefraction{1}
\def\textpagefraction{.001}

\shorttitle{Meta Generative Flow Networks with Personalization for Task-Specific Adaptation} 

\shortauthors{X. Ji et al.}  

\title [mode = title]{Meta Generative Flow Networks with Personalization for Task-Specific Adaptation}  

\tnotemark[1] 


%


\author[1,2]{Xinyuan Ji}[orcid=0000-0003-1689-0792]
\cormark[1]
\credit{Methodology, Coding, Writing - original draft}
\ead{jixinyuan1996@stu.xjtu.edu.cn}
\author[3]{Xu Zhang}[orcid=0000-0002-1629-2103]
\credit{Theoretical analysis, Supervision, Writing - original draft, and Funding acquisition}
\cormark[1]
\ead{xuzhang_cas@lsec.cc.ac.cn}

\author[1]{Wei Xi}
\credit{Supervision and Funding acquisition}
\ead{weixi.cs@gmail}
\author[4]{Haozhi Wang}
\credit{Reviewing}
\ead{wanghaozhi@tju.edu.cn}
\author[2]{Olga Gadyatskaya}
\credit{Reviewing}
\ead{o.gadyatskaya@liacs.leidenuniv.nl}
\author[5]{Yinchuan Li}[orcid=0000-0002-4263-5130]
\credit{Propose the idea, Conceptualization, Supervision, and Reviewing}
\cormark[2]
\ead{liyinchuan@huawei.com}

\affiliation[1]{organization={Faculty of Electronic and Information Engineering, Xi'an Jiaotong University},
            country={China}}

\affiliation[2]{organization={Leiden Institute of Advanced Computer Science (LIACS), Leiden University},
            country={Netherlands}}
\affiliation[3]{organization={LSEC, Academy of Mathematics and Systems Science, Chinese Academy of Sciences}, 
country={China}}

\affiliation[4]{organization={School of Electrical and Information Engineering, Tianjin University}, 
country={China}}

\affiliation[5]{organization={Noah’s Ark Lab, Huawei Technologies}, 
country={China}}


\cortext[1]{Equal contribution.}
\cortext[2]{Corresponding author.}

\begin{abstract} 
 Multi-task reinforcement learning and meta-reinforcement learning have been developed to quickly adapt to new tasks, but they tend to focus on tasks with higher rewards and more frequent occurrences, leading to poor performance on tasks with sparse rewards. To address this issue, GFlowNets can be integrated into meta-learning algorithms (GFlowMeta) by leveraging the advantages of GFlowNets on tasks with sparse rewards. However, GFlowMeta suffers from performance degradation when encountering heterogeneous transitions from distinct tasks. To overcome this challenge, this paper proposes a personalized approach named pGFlowMeta, which combines task-specific personalized policies with a meta policy. Each personalized policy balances the loss on its personalized task and the difference from the meta policy, while the meta policy aims to minimize the average loss of all tasks. The theoretical analysis shows that the algorithm converges at a sublinear rate. Extensive experiments demonstrate that the proposed algorithm outperforms state-of-the-art reinforcement learning algorithms in discrete environments.
\end{abstract}
\maketitle
\section{Introduction}
\label{intro}
 Reinforcement learning (RL) methods aim to find a policy that maximizes expected returns through interactions with the environment \cite{sutton1988learning, watkins1992q}. While RL has been successfully applied in various fields \cite{mnih2013playing, silver2016mastering, heaton2016deep, baucum2023optimizing, liang2018deep, kumari2021reinforcement}, it often struggles to adapt to new tasks, leading to suboptimal performance. 
 This is because the optimal policy for a new task may differ significantly from that of a previously learned task. Moreover, RL methods typically require a large number of interactions with the environment to learn an effective policy, which can be time-consuming and impractical in certain scenarios.

To adapt to new tasks quickly with minimal additional training, Multi-task reinforcement learning (Multi-task RL) and meta-reinforcement learning (meta-RL) have been developed. Early research on Multi-task RL is based on transfer learning \cite{weiss1999multiagent, boutsioukis2012transfer}, which transfers knowledge between related sources and target tasks to improve the performance of policy used in the target task. Subsequent works proposed common representation \cite{rosenbaum2017routing, d2020sharing, vuong2019sharing, yang2020multi}, gradients similarity \cite{yu2020gradient, chen2018gradnorm}, and policy distillation \cite{xu2020knowledge, parisotto2015actor, teh2017distral}. Additionally, Meta-RL methods such as MAML \cite{finn2017model}, E-MAML \cite{stadie2018some}, and PEARL \cite{rakelly2019efficient} aim to enable the model to acquire a "learning to learn" ability by leveraging existing knowledge to quickly learn new tasks. However, all of these methods tend to focus on tasks with higher rewards and more frequent occurrences, ignoring tasks with fewer rewards and less frequent occurrences. Consequently, the model may perform poorly on tasks with sparse rewards \cite{vithayathil2020survey, zhang2021survey}.

\begin{figure}
    \centering
\includegraphics[width=0.45\textwidth]{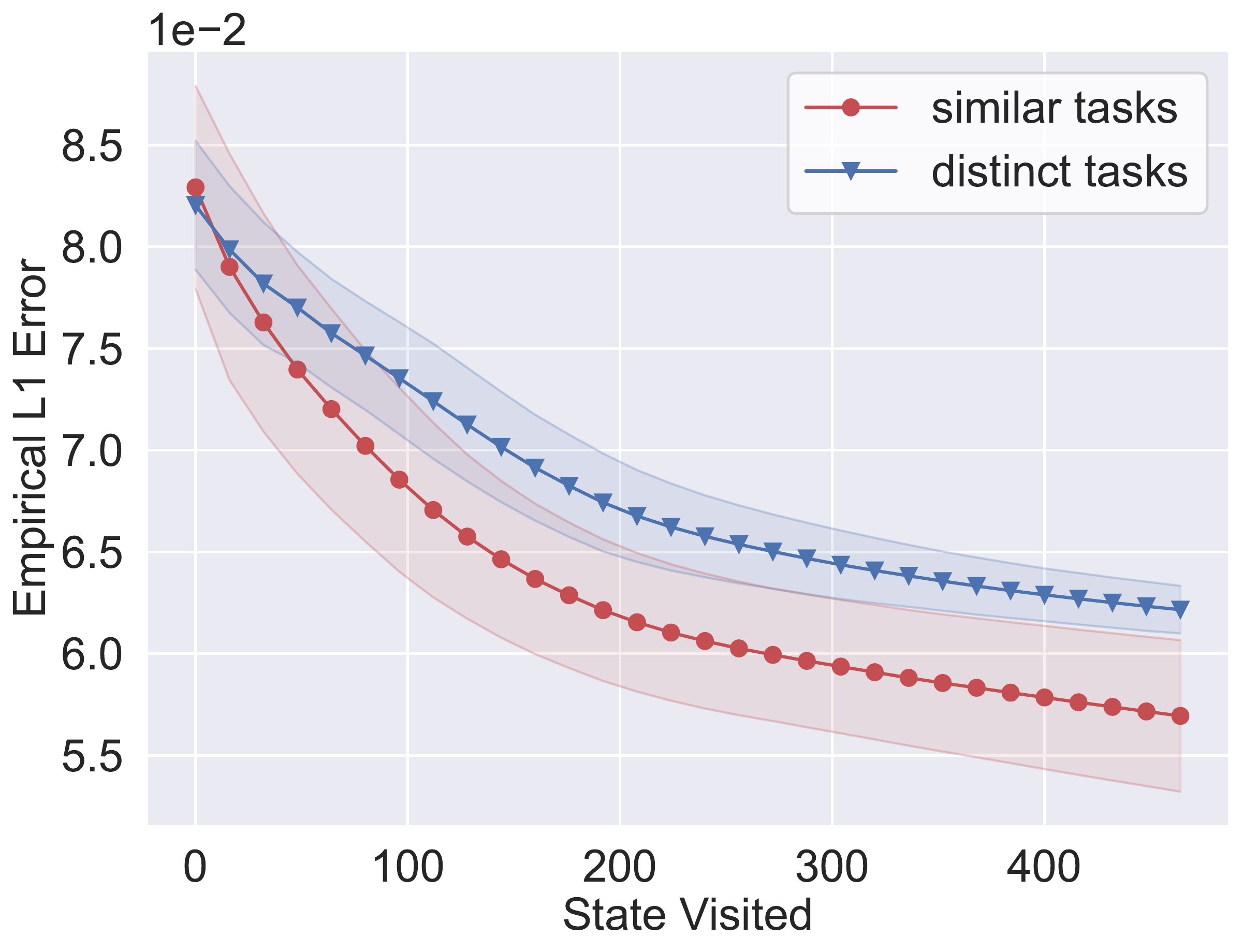}
    \caption{Averaged Empirical $L_1$ Error of GflowMeta on similar and distinct tasks in ``Frozen lake'' environment.}
    \label{fig:Motivation}
\end{figure}

\begin{sloppypar}To adapt to new tasks quickly and improve the performance on tasks with sparse rewards, a possible solution is to integrate Generative Flow Networks (GFlowNets) \cite{bengio2021flow,lahlou2023theory, jain2023gflownets, DBLP:journals/corr/abs-2111-09266} into  meta-learning, leading to a variant called GFlowMeta. 
Compared with existing works, GFlowMeta can combine the advantages of GFlowNets and Multi-task RL. It can find multiple objects within the limitation return function and adapt to unseen tasks. However, it is important to note that GFlowMeta demonstrates significant performance improvements primarily when the trained tasks are similar. Unfortunately, in cases where the tasks are distinct, the performance improvements achieved by GFlowMeta are relatively limited. In Fig. \ref{fig:Motivation}, we present a performance comparison of training with GFlowMeta on similar and distinct tasks across “Frozen Lake” environments. The example of distinct tasks can be found in Fig. \ref{distinct tasks}. In contrast to distinct tasks, similar tasks are supposed from the same domain, (i.e. the same transition distribution). The results clearly demonstrate a disparity in performance between the two scenarios.
\end{sloppypar}


\begin{figure*}[h]
\centering
\includegraphics[width=0.8\linewidth]{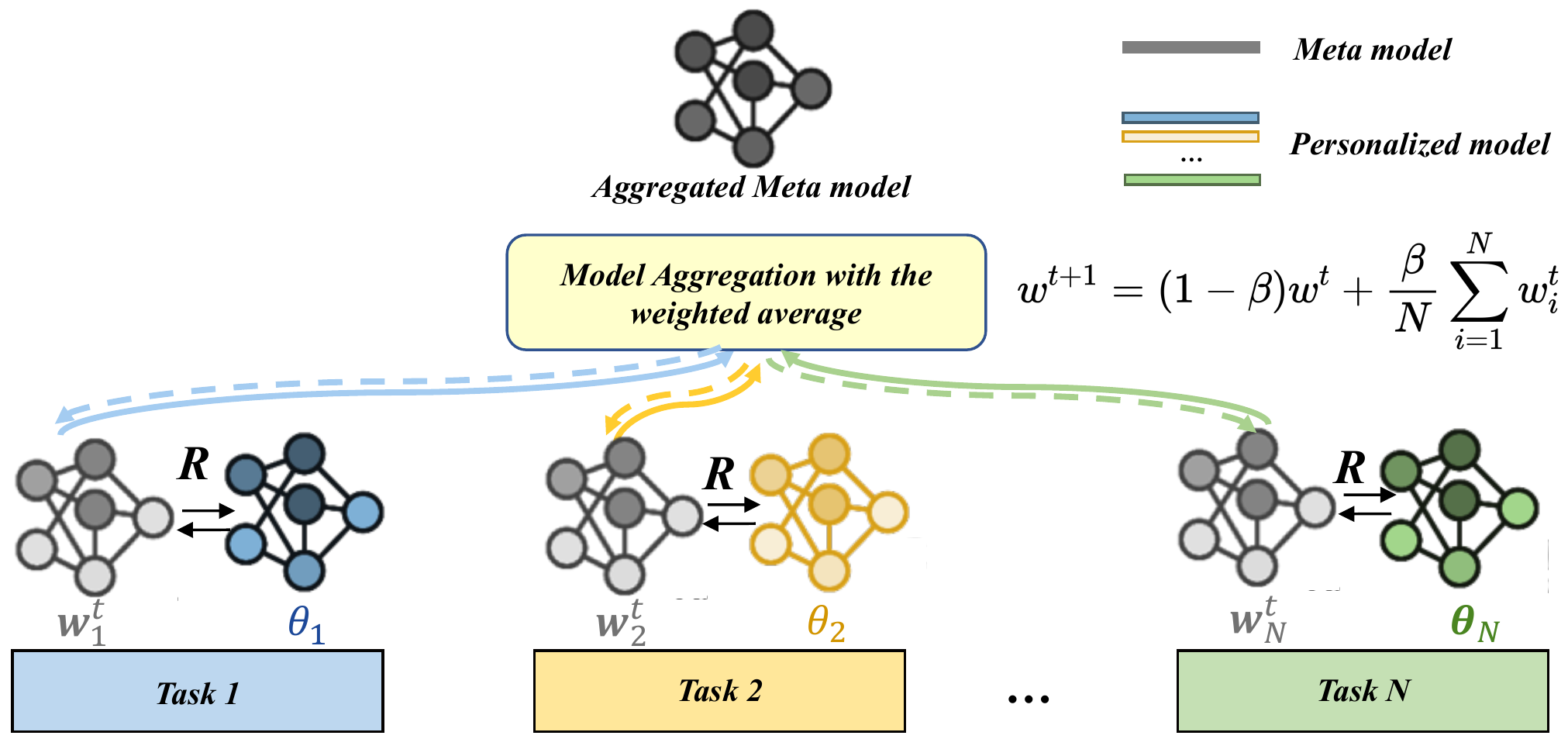}
\caption{System diagram for \pGFlowMeta. Consider a meta-learning problem in which there are $N$ tasks. In each task, the agent learns two policy models: the Personalized policy $\theta_i$ and the Meta policy $w$. The training process involves alternating updates between the Personalized policy and Meta policy models for a specified number of iterations ($R$). Following the updates, the Meta policy models are aggregated using the formula $w^{t+1}=(1-\beta)w^t+\frac{\beta}{N}\sum_{i=1}^N w_i^t$, where $\beta$ is a weighting parameter, and $N$ represents the number of tasks. This framework enables the agent to learn both personalized policies for individual tasks and a meta-level policy that captures common knowledge across tasks. 
}
\label{fig3}
\end{figure*}

In this paper, we propose a framework, as shown in Fig. \ref{fig3}, called Personalized Meta Generative Flow Networks (\pGFlowMeta) to address the challenge posed by distinct tasks. We formulate a bilevel optimization problem to optimize two policies simultaneously: the Meta Policy (MP), which aims to learn a generalized policy that can quickly adapt to new tasks, and the Personalized Policy(PP), which focuses on improving the performance over each distinct task. To achieve this, we introduce a proximal operator in the lower-level objective function to penalize the difference between the meta policy and the personalized policy. This allows us to balance the model aggregation on all tasks while also improving the performance of each task individually.  

\subsection{Main Contributions}
Our main contributions are threefold: First, we propose a personalized framework named \pGFlowMeta\ to improve the performance of \GFlowMeta\ on distinct tasks, which learns a \emph{meta policy} for all tasks and \emph{personalized policies} for specific tasks, respectively. In particular, meta policy seeks to minimize the sum of all tasks by aggregating  personalized policies while each personalized policy aims to balance the loss on its own task and the difference from the meta policy.

\begin{sloppypar}Second, an alternating minimization algorithm for \pGFlowMeta\ is proposed to achieve personalization on distinct tasks. The personalized policies are updated by solving the corresponding personalized lower-level problems inexactly while the auxiliary policies used to aggregate the meta policy are updated by using the gradient descent algorithm. Furthermore, a convergence analysis is established for the proposed algorithm, which proves that the algorithm converges sublinearly as the number of iterations increases. Besides, we provide the convergence of personalized policies to meta policy: the personalized policies on average converge to a neighborhood of the meta policy with a sublinear rate. Moreover, to understand how different hyperparameters affect the convergence of \pGFlowMeta, we conduct various experiments on the “Grid World” environment.\end{sloppypar} 

Finally, we make numerical experiments to compare the performance of \pGFlowMeta\ with other well-known algorithms on three discrete reinforcement environments: “Grid World”, “Frozen Lake”, and “Cliff Walking”. The experimental results show that the proposed \pGFlowMeta\ can achieve better performance than GflowMeta on distinct tasks. Compared to other previous reinforcement learning algorithms, the proposed \pGFlowMeta\ algorithm can achieve the highest average rewards, obtain the lowest empirical $L_1$ error, and find the largest number of modes with the greatest likelihood in the three environments. Moreover, to understand how different hyperparameters affect the convergence of \pGFlowMeta, we also conduct various experiments on the “Grid World” environment.

\subsection{Related Works}

This section introduces some works that are related to our research including meta-reinforcement learning, multi-task learning, and GFlowNets.

\vspace{0.1cm}
\noindent\textbf{Meta-reinforcement Learning.}
Meta-reinforcement learning (meta-RL) is a subfield of reinforcement learning (RL) that focuses on learning how to learn in the context of RL \cite{kaelbling1996reinforcement, duan2016benchmarking, sutton2018reinforcement}. Specifically, meta-RL aims to develop algorithms and models that can quickly adapt to new tasks and environments by leveraging knowledge learned from previous tasks\cite{finn2017model, rakelly2019efficient}.
Different approaches have been proposed in the meta-RL literature, including model-free and model-based methods. Model-free methods such as RL$^2$, MAML, ProMP, E-MAML, MAESN, ES-MAML, et al. \cite{duan2016rl, finn2017model, rothfuss2018promp, stadie2018some, gupta2018meta, song2019maml} directly learn a policy that can adapt to new tasks without explicitly modeling the environment. Model-based methods, on the other hand, learn a model of the environment and use it to simulate different scenarios and generate training data for the meta policy. For instance, PEARL \cite{rakelly2019efficient} and CaDM \cite{lee2020context} learn a neural network with task latent variables to represent a meta model of state transitions and a task inference network. They plan agent policies by inferring task information and adapting the state transition model accordingly. The aforementioned methods achieve fast adaptation to new tasks by training on multiple similar tasks. However, they often encounter model divergence when applied to tasks with low similarity due to environmental heterogeneity.

\vspace{0.1cm}
\noindent\textbf{Multi-task Learning $\&$ Multi-task Reinforcement Learning.}
Multi-Task Learning (MTL) is a machine learning approach that aims to improve the performance of a model by jointly learning multiple related tasks \cite{caruana1997multitask, ruder2017overview, zhang2018overview, zhang2021survey}. In the context of reinforcement learning, leveraging MTL is one direction to enhance the efficiency of RL agents. By simultaneously training the agent to perform multiple related tasks, it can potentially learn to generalize better and more efficiently than a single-task agent. Several approaches and techniques have been developed to improve the effectiveness of RL agents, with an increasing focus on using Multi-task RL. Much of the research in this area, such as that by \cite{weiss1999multiagent, boutsioukis2012transfer, rusu2016progressive, PathNet} has centered around transfer learning or continual learning, which aims to transfer knowledge between related source and target tasks to improve the performance of machine learning (ML) algorithms for the target task. For example, Policy Distillation \cite{ teh2017distral, xu2020knowledge} and Actor-Mimic \cite{parisotto2015actor} leverage the concept of distillation to achieve multi-task deep reinforcement learning. The key idea is to compress the features learned by a complex model into a smaller and faster model.

\noindent\textbf{Generative Flow Networks.} Generative Flow Networks (GFlowNets) were initially introduced by Bengio in 2021 \cite{bengio2021flow} as a generative modeling framework for sampling from a distribution. Since then, GFlowNets have been extensively studied \cite{nica2022evaluating, li2023cflownets, zhang2022unifying, jain2022biological, madan2022learning, jain2023gflownets}. Madan et al. \cite{madan2022learning} addressed the bias-variance tradeoff challenge by enabling the training of GFlowNets in environments with longer action sequences and sparser reward landscapes. 
Jain et al. \cite{jain2022biological} proposed an active learning algorithm that combines epistemic uncertainty estimation and GFlowNets to generate diverse and useful candidate solutions for designing biological sequences. 
Zhang et al. \cite{zhang2022unifying} proposed a framework that establishes connections between existing deep generative models and the GFlowNet framework, providing a unifying viewpoint through the lens of learning with Markovian trajectories. 
Li et al. \cite{li2023cflownets} addressed the challenge of handling continuous control tasks in generative modeling by proposing generative continuous flow networks (CFlowNets). 
Their work introduced a theoretical formulation, and a training framework, and demonstrated the superior exploration ability of CFlowNets compared to reinforcement learning methods through experimental results.


\section{Preliminaries}
This section introduces some important concepts in reinforcement learning and GFlowNets.

\subsection{Reinforcement learning}
{Reinforcement learning (RL) \cite{kaelbling1996reinforcement, sutton2018reinforcement} algorithms are designed to address sequential decision-making problems. In the RL, an \emph{Agent} learns the optimal \emph{Policy} through interactions with the \emph{Environment} to take \emph{Actions} on the \emph{Observation} to maximize the cumulative \emph{Reward} over a long-term horizon.} RL can be formulated as a standard Markov decision process (MDP), represented by a tuple $\mathcal{T}=\left \langle\mathcal{S}, \mathcal{A}, \mathcal{P}, \mathcal{R}, \gamma \right \rangle$, where $\mathcal{S}$ is a finite set of states, $\mathcal{A}$ is a finite set of actions, $r \in \mathcal{R}(s, a): \mathcal{S} \times \mathcal{A} \rightarrow \mathbb{R}$ is the reward function with $\mathcal{R}(s, a) =\mathbb{E}\left[R_{t+1} \mid \mathcal{S}_t=s, \mathcal{A}_t=a\right]$, $\mathcal{P}(s^{\prime}\left|s\right., a): \mathcal{S} \times \mathcal{A} \times \mathcal{S} \rightarrow \left[0,1\right]$ denotes the state transition function, and $\gamma \in \left[0,1\right)$ is the discount factor. The goal of RL is to find the optimal policy 
$\pi^{*}(a \mid s)=\mathbb{P}\left(\mathcal{A}_t=a \mid \mathcal{S}_t=s\right)$ by maximizing the expected long-term reward.   

\subsection{Generative Flow Network}
Let $X \in \mathcal{S}$ represent the set of terminal states. The primary objective of GFlowNets is to address the machine learning problem of transforming a given positive reward or return function into a generative policy $\pi(a|s) \propto R(x)$. This policy allows for the sampling of actions, as shown in Equation \ref{pi(x)}, by assigning probabilities proportional to the expected reward associated with each action
\begin{equation}
\label{pi(x)}
    \pi(x) \approx \frac{R(x)}{Z}=\frac{R(x)}{\sum_{x^{\prime} \in X} R\left(x^{\prime}\right)},
\end{equation}
where $R(x) > 0$ is a reward for a terminal state $x$. GFlowNets sees the MDP as a flow network, that is, leverages the DAG structure of the MDP. Define $s'=T(s,a)$ as the transition $(s, a)$ leads to state $s'$ and $F(s')$ as the total flow going through $s'$. For the state $s'$, it may have multiple parents, i.e. $\left|\left\{(s, a) \mid T(s, a)=s^{\prime}\right\}\right| \geq 1$, except for the root, which has no parent. Define an edge/action flow $F(s, a) = F(s \rightarrow s')$ as the flow through an edge $s \rightarrow s'$. The training process of vanilla GFlowNets needs to sum the flow of  parents and children through nodes (states), which depends on the discrete state space and discrete action space. The framework is optimized by the following flow consistency equations, where $R(s) = 0$ for interior nodes and $\mathcal{A}(s)=\varnothing$ for leaf (sink/terminal) nodes:
\begin{equation}
\label{flow_in_out}
    \sum_{s, a: T(s, a)=s^{\prime}} F(s, a)=R\left(s^{\prime}\right)+\sum_{a^{\prime} \in \mathcal{A}\left(s^{\prime}\right)} F\left(s^{\prime}, a^{\prime}\right),
\end{equation}
where $ \mathcal{A}\left(s^{\prime}\right)$ is the set of all sequences of actions allowed after state $s'$. It means that for any node $s'$, the incoming flow equals the outgoing flow, which is the total flow $F(s')$ of node $s'$. With $F$ being a flow, $F(s, a) > 0, \forall s, a$. We parameterize the unknown function $F$ with parameters $w$ and denote it as $F_w$. This could yield the following objective for a trajectory $\tau$:
\begin{align}\label{GFN}
    \mathcal{L}(w)&=\sum_{s^{\prime} \in \tau \neq s_{0}}\bigg(\sum_{s, a: T(s, a)= s^{\prime}} F_{w}(s, a) - R \left(s^{\prime}\right) \nonumber \\
    &\qquad \qquad \qquad \quad  - \sum_{a^{\prime} \in \mathcal{A}\left(s^{\prime}\right)} F_{w}\left(s^{\prime}, a^{\prime}\right) \bigg)^{2}.
\end{align}

\begin{algorithm}[tb]
\caption{GFlowMeta: A Meta Generative Flow Network} 
\begin{algorithmic}[1]
\begin{sloppypar}\STATE \textbf{Input}: training tasks $\left\{\mathcal{T}_i\right\}_{i= 1}^N$ sampled from $p\left(\mathcal{T}\right)$, $w^0$, $\eta$\end{sloppypar}
\FOR{$t = 1, \cdots, T$}
\FOR{$i=1,\ldots,N$ parallel}
\STATE  $w_{i, 1}^t=w^t$ 
\FOR{$r = 1, \cdots, R$}
\STATE Sample trajectories $\{\tau_i^k\}_{k=1}^K$ from Task $\mathcal{T}_i$
\STATE Evaluate $\nabla \mathcal{L}_i(w^{t}_{i,r})$ using  $\{\tau_i^k\}_{k=1}^K$
\STATE Compute updated parameters by solving Eq. \eqref{Meta-GFN-2} with gradient descent: $w_{i, r+1}^{t}=w_{i, r}^{t}-\eta \nabla \mathcal{L}_i(w^{t}_{i,r})$
\ENDFOR
\ENDFOR
\STATE Update meta parameters  $w^{t+1}= \frac{1}{N} \sum_{i=1}^N {w_{i,R}^t}$
\ENDFOR
\STATE \textbf{Output}: meta parameters $w^{T}$
\end{algorithmic}
\label{alg:GFlowMeta}
\end{algorithm}

\section{Personalized Meta Generative Flow Networks}

In this section, we propose a personalized meta generative flow network model (\pGFlowMeta) for distinct tasks in reinforcement learning. To make this easier to follow, we first present a meta generative flow network model (\GFlowMeta) for similar tasks. The algorithms for \GFlowMeta\ and  \pGFlowMeta\ are shown in Algorithm \ref{alg:GFlowMeta} and Algorithm \ref{alg:pGFlowMeta}, respectively.

\subsection{Meta Generative Flow Networks}

The Meta Generative Flow Network (\GFlowMeta) is a framework that focuses on learning a meta stochastic policy capable of adapting to new tasks with limited samples in reinforcement learning. The objective is to ensure that the generated objects from the policy have probabilities that are proportional to the corresponding given rewards associated with the new tasks. Let the tasks $\{\mathcal{T}_i\}$ follow the distribution $p(\mathcal{T})$, i.e., $\mathcal{T}_i \sim p(\mathcal{T})$. Assuming that $N$ tasks are sampled from the task distribution $p(\mathcal{T})$. Our goal is to learn a good policy $\pi(a|s)$  that can be easily adapted to new tasks. Since the policy in GFlowNets satisfies  $\pi(a|s)=F(s,a)/F(s)$, it's enough to learn a parameterized flow function $F_w$ that minimizes the following objective function:
\begin{equation}\label{Meta-GFN-E}
\begin{split}
    \min_{w\in \mathbb{R}^d} \left\{\mathcal{L}(w) := \frac{1}{N} \sum_{i=1}^N
     \mathcal{L}_i(w)
    \right\},
\end{split}
\end{equation}
where 
\begin{multline}\label{Meta-GFN-1}
    \mathcal{L}_{i}(w)=\mathbb{E}_{\tau \sim p_i(\tau)}  \Bigg[\sum_{s^{\prime} \in \tau \neq s_{0}}\bigg(\sum_{s, a: T(s, a)= s^{\prime}} F_{w}(s, a) - R \left(s^{\prime}\right) \\
     - \sum_{a^{\prime} \in \mathcal{A}\left(s^{\prime}\right)} F_{w}\left(s^{\prime}, a^{\prime}\right) \bigg)^{2}\Bigg].
\end{multline}
Here, $w \in \mathbb{R}^d$ denotes the parameter of the flow function  $F_w$ to be learned, and $p_i(\tau)$ denotes the trajectory distribution of the task $\mathcal{T}_i$.

Assuming $K$ trajectories are sampled from each trajectory distribution ${p_i(\tau)}$, then the objective function becomes 
\begin{equation}\label{Meta-GFN-S}
\begin{split}
    \min_{w\in \mathbb{R}^d} \left\{\mathcal{L}(w) := 
    \frac{1}{N} \sum_{i=1}^N  \tilde{\mathcal{L}}_i(w) 
    \right\},
\end{split}
\end{equation}
where 
\begin{align}\label{Meta-GFN-2}
    \tilde{\mathcal{L}}_{i}(w)&=\frac{1}{K}  \sum_{k=1}^K \Bigg[\sum_{s^{\prime}_i \in \tau_i^k\neq s_{i,0}}\bigg(\sum_{s, a: T(s_i, a_i)= s^{\prime}_i} F_{w}(s_i, a_i)  \nonumber \\
    &\qquad \qquad - R \left(s^{\prime}_i\right)  - \sum_{a^{\prime}_i \in \mathcal{A}\left(s^{\prime}_i\right)} F_{w}\left(s^{\prime}_i, a^{\prime}_i\right) \bigg)^{2}\Bigg].
\end{align}
Here, $\tau_i^j$ denotes the $j$-th trajectory of the task $\mathcal{T}_i$.

For non-convex objectives \eqref{Meta-GFN-E} and \eqref{Meta-GFN-1}, GFlowMeta has the potential to learn a shared prior over model parameters, resulting in a global model by averaging models in the parameter space based on the tasks. The complete pseudo-code for \GFlowMeta\ can be found in Algorithm \ref{alg:GFlowMeta}. The training process of \GFlowMeta\ primarily consists of two loops: an outer loop for the aggregation of the meta policy model and an inner loop for the updates of the meta policy model for each task. In the $t$-th round of the outer loop, every task $\mathcal{T}_i$ samples trajectories $\{\tau_i^k\}_{k=1}^K$ and the meta policy model is updated in the inner loop based on $w^t_{i, r}$ to optimize objective \eqref{Meta-GFN-2}. The updated meta policy model after the $r$-th update in the $t$-th round is denoted as $w^t_{i,r+1}$. Once all tasks have completed the inner loop training procedure, the new meta policy model $w^{t+1}$ is obtained by aggregating the inner updates of the tasks through averaging. 


While \GFlowMeta\ can be effective for similar tasks, it may encounter significant performance degradation when applied to tasks that are not closely related. For instance, if the tasks involve disparate datasets or exhibit different input features, \GFlowMeta\ may struggle to capture the underlying data structure and yield subpar performance. Similarly, if the tasks involve distinct control dynamics, GFlowMeta may struggle to generalize to new tasks with different dynamics. As a result, GFlowMeta might not be suitable for different tasks. It is crucial to carefully consider the task characteristics and assess the compatibility between GFlowMeta and the specific task requirements before applying this approach.



\subsection{Personalized Meta Generative Flow Networks}

To deal with distinct tasks, we propose a personalized meta Generative Flow Network (\pGFlowMeta), where we learn a meta policy that can be adapted to multiple distinct tasks and learn a personalized policy for each task. Assuming that $N$ distinct tasks are sampled from the task distribution $p(\mathcal{T})$. The goal of \pGFlowMeta\ is to optimize the following objective function
\begin{equation}\label{pMeta-GFN}
\begin{split}
    \min_{w\in \mathbb{R}^d} \left\{\mathcal{L}(w) := \frac{1}{N} \sum_{i=1}^N \mathcal{L}_i(w)
    \right\},
\end{split}
\end{equation}
where
\begin{equation}\label{pMeta-GFN-1}
\begin{split}
    \mathcal{L}_i(w) = \min_{\theta_i}  \ell_i(\theta_i)  + \lambda f(w, \theta_i)
\end{split}
\end{equation}
and
\begin{equation}\label{pMeta-GFN-2}
\begin{split}
    \ell_{i}(\theta_i)=\mathbb{E}_{\tau_i \sim p_i(\tau)} \Bigg[ \sum_{s_i' \in \tau_i \neq s_{0}}\bigg(\sum_{s_i, a_i: T_i(s_i, a_i)= s_i'} F_{\theta_i}(s_i, a_i) \\ - R (s_i') 
    - \sum_{a_i' \in \mathcal{A}(s_i')} F_{\theta_i}\left(s_i', a_i'\right) \bigg)^{2} \Bigg].
\end{split}
\end{equation}
Here, $f(w,\theta_i)=\frac{1}{2}\|w-\theta_i\|^2$ is a regularization term that penalizes the difference between $w$ and $\theta_i$, $w \in \mathbb{R}^d$ is the parameter of the meta flow function $F_{w}$ and $\theta_i$ is the parameter of the personalized flow function $F_{\theta_i}$. Let $\theta_i^\star$ be the solution of \eqref{pMeta-GFN-1}, i.e.,
\begin{equation}\label{pMeta-GFN-s}
\begin{split}
    \theta_i^\star(w) = \arg \min_{\theta_i \in \mathbb{R}^d}  \ell_i(\theta_i)  + \lambda f(w, \theta_i).
\end{split}
\end{equation}

Assuming $K$ trajectories are sampled from each trajectory distribution ${p_i(\tau)}$, then the objective function in  \eqref{pMeta-GFN-1}  becomes 
\begin{equation}\label{pMeta-GFN-a}
\begin{split}
   \tilde{\mathcal{L}}_i(w) = \min_{\theta_i}  \tilde{\ell}_i(\theta_i)  + \lambda f(w, \theta_i)
\end{split}
\end{equation}
and
\begin{equation}\label{pMeta-GFN-b}
\begin{split}
    \tilde{\ell}_{i}(\theta_i)= \frac{1}{K} \sum_{k=1}^K \Bigg[ \sum_{s_i' \in \tau_i^k \neq s_{0}}\bigg(\sum_{s_i, a_i: T_i(s_i, a_i)= s_i'} F_{\theta_i}(s_i, a_i) \\ - R (s_i') 
    - \sum_{a_i' \in \mathcal{A}(s_i')} F_{\theta_i}\left(s_i', a_i'\right) \bigg)^{2} \Bigg].
\end{split}
\end{equation}

Define $\rho_w^i(\theta_i)=\ell_i(\theta_i)  + \lambda f(w, \theta_i)$ and $\tilde{\rho}_w^i(\theta_i)=\tilde{\ell}_i(\theta_i)  + \lambda f(w, \theta_i)$. Let $\hat{\theta}_i(w)$ be the inexact solution of Eq. \eqref{pMeta-GFN-a}
\begin{equation}\label{pMeta-GFN-t}
\begin{split}
    \hat{\theta}_i(w) \approx \arg \min_{\theta_i \in \mathbb{R}^d}  \tilde{\rho}_w^i(\theta_i).
\end{split}
\end{equation}
such that $\hat{\theta}_i(w)$ is the $\zeta$-approximation of $\theta_i^\star(w)$, i.e., $\mathbb{E}\big[\|\hat{\theta}_i(w)-\theta_i^\star(w)\|^2 \big]\le \zeta^2$. The gradient of $\tilde{\mathcal{L}}_i(w)$ with respect to $w$ used for the updates of meta policy model is 
\begin{equation} \label{eq: Gradient_L}
    \nabla \tilde{\mathcal{L}}_i(w) = \lambda (w-\hat{\theta_i}(w)).
\end{equation}

\begin{algorithm}[tb]
\caption{\pGFlowMeta: A Personalized Meta Generative Flow Network} 
\begin{algorithmic}[1] 
\STATE \textbf{Input}: training tasks $\left\{\mathcal{T}_i\right\}_{i=1}^{N}$ sampled from $p\left(\mathcal{T}\right)$, $w^0$, $\eta$, $\beta$
\FOR{$t = 1, \cdots, T$}

\FOR{$i=1,\ldots,N$ parallel}
\STATE  $w_{i,1}^t=w^t$ 
\FOR{$r = 1, \cdots, R$}
\STATE Sample trajectories $\{\tau_i^k\}_{k=1}^K$ from Task $\mathcal{T}_i$
\STATE Evaluate $\nabla \ell_i(\theta_i)$ using  $\{\tau_i^k\}_{k=1}^K$
\STATE Update personalized parameters by solving Eq. \eqref{pMeta-GFN-t} inexactly to obtain a $\zeta$-approximate $\hat{\theta}_{i}(w_{i,r}^t)$ \\
\STATE Update auxiliary meta parameters \\$w_{i,r+1}^t=w_{i,r}^t-\eta \lambda(w_{i,r}^t-\hat{\theta}_{i}(w_{i,r}^t)) $
\ENDFOR
\ENDFOR
\STATE Update meta parameters  $w^{t+1}=(1-\beta) w^t + \frac{\beta}{N} \sum_{i=1}^N {w_{i,R}^t}$
\ENDFOR
\STATE \textbf{Output}: meta parameters $w^{T}$ and personalized parameters $\{ \theta_{i,R}^T\}_{i=1}^N $
\end{algorithmic}
\label{alg:pGFlowMeta}
\end{algorithm}

As illustrated in Algorithm \ref{alg:pGFlowMeta}, the training process of \pGFlowMeta\ consists of two main steps. In the outer loop, the meta policy is aggregated, while in the inner loop, both the auxiliary meta policy model and the personalized policy are updated for each task. In the $r$-th inner update of the $t$-th round, every task $\mathcal{T}_i$ samples trajectories $\{\tau_i^k\}_{k=1}^K$ and firstly updates the personalized policy by solving \eqref{pMeta-GFN-t} inexactly. 
Then the auxiliary meta policy model is updated by using gradient descent, i.e., 
\begin{equation} \label{eq: wt_ir_update_2}
    w_{i,r+1}^t=w_{i, r}^{t}-\eta \nabla \mathcal{L}_i(w^{t}_{i,r})=w_{i,r}^t-\eta \lambda(w_{i,r}^t-\hat{\theta}_{i}(w_{i,r}^t)).
\end{equation}
After $R$ rounds of inner updates, the new meta policy model $w^{t+1}$ is aggregated by $w^{t+1}=(1-\beta) w^t + \frac{\beta}{N} \sum_{i=1}^N {w_{i,R}^t}$, where $\beta>0$ is introduced to accelerate the convergence of the algorithm. These steps are executed repeatedly until reaching the $T$th round.

For ease of analysis, we rewrite the updates with concise symbols 
\begin{align}
    w_{i,r+1}^t&=w_{i,r}^t-\eta \lambda(w_{i,r}^t-\hat{\theta}_{i}(w_{i,r}^t))= w_{i,r}^t-\eta \lambda g_{i,r}^t  \label{eq: wt_ir_update}\\
    w^{t+1}&=(1-\beta)w^t +\frac{\beta}{N} \sum_{i=1}^N w^t_{i,R}\nonumber\\
    &=w^t-\eta \beta R \frac{1}{NR} \sum_{i=1}^N \sum_{r=1}^R (w_{i,r}^t-\hat{\theta}_{i}(w_{i,r}^t)) \nonumber\\
    &\triangleq w^t- \tilde{\eta} g^t, \label{eq: wt_update}
\end{align}
\begin{sloppypar} where $g_{i,r}^t=w_{i,r}^t-\hat{\theta}_{i}(w_{i,r}^t)$, $\tilde{\eta} \triangleq \eta \beta R$ and $g^t \triangleq \frac{1}{NR} \sum_{i=1}^N \sum_{r=1}^R (w_{i,r}^t-\hat{\theta}_{i}(w_{i,r}^t) \triangleq \frac{1}{NR} \sum_{i,r=1}^{N,R} g_{i,r}^t$. Here, $\tilde{\eta}$ and $g^t$ can be regarded as the step size and the approximate gradient of the outer loop, respectively.\end{sloppypar}

\section{Theoretical analysis}

In this section, we provide the convergence guarantee for the proposed personalized approach pGFlowMeta. The proofs are presented in the Appendices. To facilitate the following analysis, we make some reasonable assumptions.

\begin{assumption} \label{assumption_1_variance}
The variance of stochastic gradients of each task is bounded 
\begin{align} \label{assumption_1_variance_sg}
    \mathbb{E}\Big[ \norm{ \nabla \ell_i(\theta_i) - \nabla \tilde{\ell}_{i}(\theta_i)} ^2\Big] \le \kappa_1^2.
\end{align}
The variance of each local gradient to the average of all local gradients is bounded
\begin{align} \label{assumption_1_variance_l_g}
    \bigg{\|} \nabla \ell_i(w) - \frac{1}{N} \sum_{j=1}^N \nabla \ell_j(w)  \bigg{\|}^2 \le \kappa_2^2.
\end{align}
\end{assumption}
\begin{remark}
    Since we know $\mathbb{E}[ \tilde{\ell}_i(\theta_i)] = \ell_i(\theta_i)$ for $K=1$, and the variance $\mathbb{E}[ \| \nabla \ell_i(\theta_i) - \nabla \tilde{\ell}_{i}(\theta_i)\| ^2]$ decreases with the increase of $K$, the assumption \eqref{assumption_1_variance_sg} can be easily satisfied for arbitrary $\kappa_1$ with a sufficiently large $K$. The assumption \eqref{assumption_1_variance_l_g} is standard to bound the diversity of all tasks.
\end{remark}

\begin{assumption} \label{assum:F_theta}
The functions $\{F_{\theta_i}\}_{i=1}^{N}$ are $B$-bounded, $L_0$-Lipschitz continuous and $L_1$-Lipschitz gradients, i.e.,
\begin{align}
    \|F_{\theta_i}\| &\le B,\label{assumption_2_bound}\\
     \|F_{\theta_i}-F_{\theta_i^\prime}\| &\le L_0 \|\theta_i-\theta_i^\prime\|, \label{assumption_2_Lipschitz}\\
      \|\nabla F_{\theta_i}-\nabla F_{\theta_i^\prime}\| &\le L_1 \|\theta_i-\theta_i^\prime\|. \label{assumption_2_gradient_Lipschitz}
\end{align}
\end{assumption}

\begin{remark}
The assumption \eqref{assumption_2_bound} is due to the flow consistency equations \eqref{pi(x)} and the boundness of the rewards. The assumption \eqref{assumption_2_Lipschitz} and \eqref{assumption_2_gradient_Lipschitz} are assumed to guarantee the function $F_{\theta_i}$ is Lipschitz continuous and smooth, which limits the function to not fluctuate too much.
\end{remark}

Considering reinforcement learning environments with discrete action spaces, we assume the states and rewards are limited as follows.

\begin{assumption} \label{assum: limited state} The states in the GFlowNets are finite. The maximum length of the trajectories sampled from $\{\mathcal{T}_i\}_{i=1}^N$ is $H_1$ and the maximum number of parent nodes and child nodes for all states is $H_2$, where $0<H_1, H_2<\infty$ are positive integers.
\end{assumption}

\begin{assumption} \label{assum: reward}
 The rewards are finite with a maximum $H_0$.
\end{assumption}

Next, we present some important properties for the lower-level and upper-level objective functions, which show their smoothness respectively.

\begin{lemma} \label{lm: smooth_L_l}
Assuming the Assumptions \ref{assum:F_theta}, \ref{assum: limited state} and \ref{assum: reward} are true, then $\ell_{i}(\cdot)$ is $L_\ell$-smooth, where 
\begin{equation}
    L_\ell=H_1\Big[(H_0+2H_2 B)2 H_2 L_1+ 4H_2^2 L_0^2\Big].
\end{equation}
\end{lemma}

\begin{lemma} \label{lm: smooth_L_L}
If $\lambda>2 L_{\ell}$, then the function ${\mathcal{L}}$ and ${\mathcal{L}}_i$ are $L_\mathcal{L}$-smooth with $L_\mathcal{L}\triangleq \lambda$.    
\end{lemma}

This following theorem shows that we can get the $\zeta$-approximation of $\theta_i^\star(w)$ once $\hat{\theta}_i(w)$ satisfy $\|\nabla \tilde{\rho}_w^i(\hat{\theta}_i(w)) \| \le \delta$, which can serves as a stopping criterion of solving \eqref{pMeta-GFN-t} inexactly.

\begin{theorem} \label{thm: inexactsolution}
    Let $\theta_i^\star(w)$ be the solution of \eqref{pMeta-GFN-s}, and $\hat{\theta}_i(w)$ satisfy $\|\nabla \tilde{\rho}_w^i(\hat{\theta}_i(w)) \| \le \delta,$ if $\lambda >L_{\ell}$, then we have 
\begin{align}
    \mathbb{E}\big[\|\hat{\theta}_i(w)-\theta_i^\star(w)\|^2 \big]\le \zeta^2 \triangleq \frac{2 ( \kappa_1^2 +  \delta^2 )}{(\lambda - L_\ell)^2} .
\end{align}
\end{theorem}

Finally, we provide the convergence analysis of the proposed pGFlowMeta algorithm. 

\begin{theorem} \label{thm: convergence} Suppose that Assumpsions \ref{assumption_1_variance}-\ref{assum: reward} are true. Let the step size satisfy $\eta \le \min \Big\{\frac{\tilde{\eta}_o}{\beta R}, \frac{1}{2L_{\mathcal{L}}\sqrt{(1+R)R}} \Big\}$, where $\tilde{\eta}_o \triangleq \frac{1}{70L_{\mathcal{L}} \lambda^2}$ and $\lambda^2-8L_{\ell}^2\ge 1$.
Let $\tilde{t}$ be sampled uniformly from the index set $\{1,\ldots,T\}$. Then the following inequalities hold
\begin{multline}
      \mathbb{E} [ \| \nabla \mathcal{L}(w^{\tilde{t}})\|^2] 
    \\ \le  \mathcal{O} \bigg(\frac{\Delta}{\tilde{\eta}_o T} + \sqrt[3]{\frac{\Delta^2 L_{\mathcal{L}}^2 (\lambda^2 \zeta^2 + \kappa_{\mathcal{L}}^2) }{\beta^2 T^2}} + \sqrt{\frac{\Delta L_{\mathcal{L}}   \kappa_{\mathcal{L}}^2}{T}} + \lambda^2 \zeta^2\bigg),
\end{multline}
and
\begin{multline}
      \frac{1}{N} \sum_{i=1}^N \mathbb{E}\Big[\|\hat{\theta}_i(w_{\tilde{t}})-w_{\tilde{t}}\|^2 \Big]  
    \\ \le  \mathcal{O} \bigg(\zeta^2 + \frac{\kappa_{\mathcal{L}}^2}{\lambda^2} \bigg) + \mathcal{O} \Big({\mathbb{E} \big[ \| \nabla \mathcal{L}(w^{\tilde{t}})\|^2 \big]} \Big),
\end{multline}
where  $\kappa_{\mathcal{L}}^2=\frac{{\lambda^2} \kappa_2^2}{{\lambda^2}-{8L_{\ell}^2}}$, $\Delta \triangleq \mathcal{L}(w^{1})-\mathcal{L}(w^\star)$ and $w^\star$ is the optimal solution of $\mathcal{L}(w)$.
\end{theorem}
\begin{remark} Theorem \ref{thm: convergence} demonstrates that the algorithm converges with a sublinear convergence rate $1/\sqrt{T}$. Besides, the personalized policies converge in average to a neighborhood of the meta policy with a sublinear rate. The radius of the neighborhood is determined by the regularization parameter $\lambda$, the approximation error $\zeta$, and the parameter $\kappa_{\mathcal{L}}$.
\end{remark}

\begin{figure*}[t]
\centering  
\subfigure[]{   
\begin{minipage}{6cm}
\centering    
\includegraphics[width=1\linewidth]{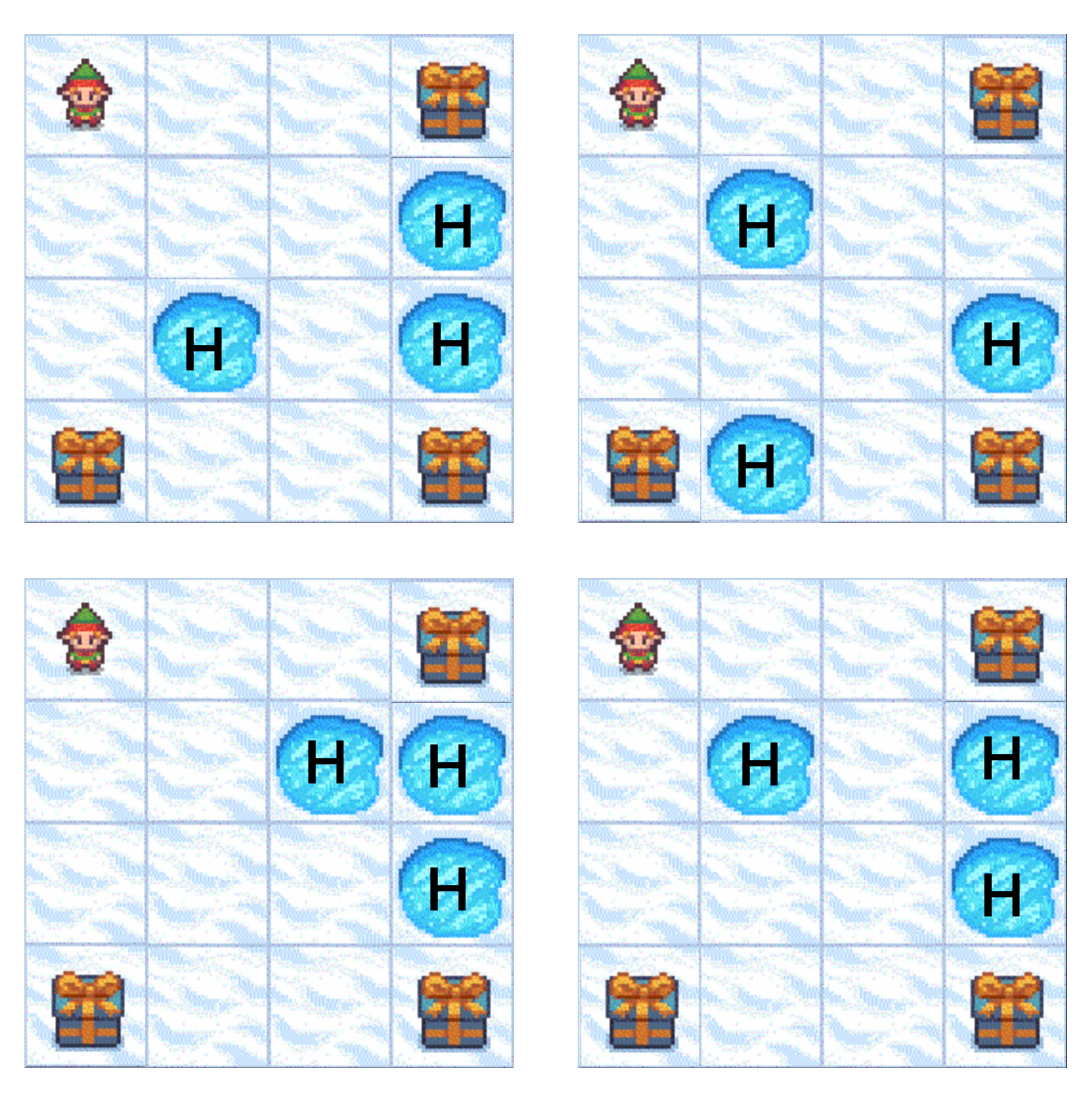}  
\end{minipage}
}
\subfigure[]{ 
\begin{minipage}{9.2cm}
\centering    
\includegraphics[width=1\linewidth]{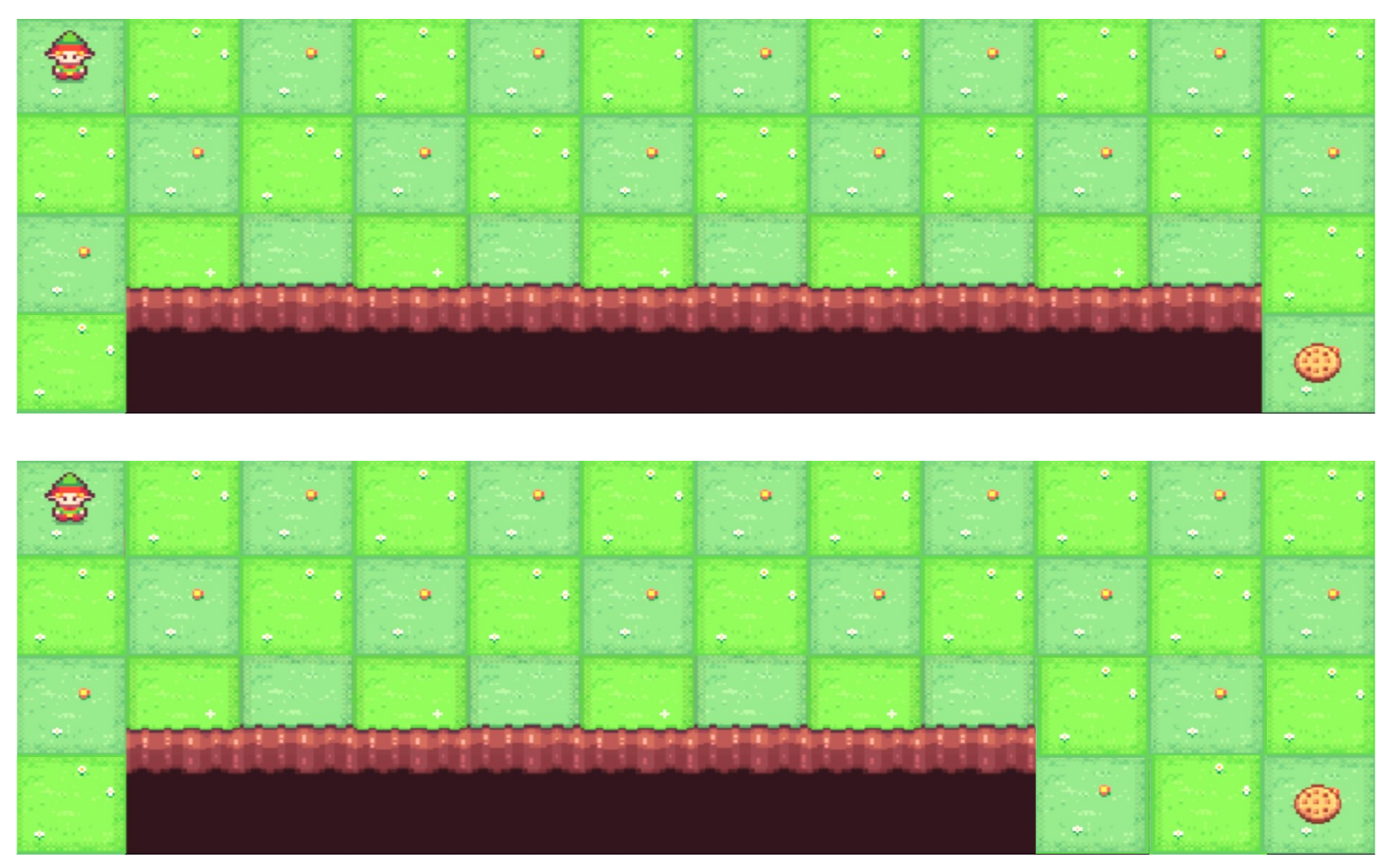}
\end{minipage}
}
\caption{Two examples illustrating the generation of distinct tasks. (a) Distinct tasks in the “Frozen Lake” environment. These tasks were characterized by variations in the location of holes. Different tasks may have holes placed at different positions, resulting in varying levels of difficulty. (b) Distinct tasks in the “Cliff Walking” environment. These tasks were characterized by variations in the length of the cliff. Each task may have a different cliff length, leading to varying levels of risk and difficulty. On distinct tasks, the agent's transitions would differ across tasks despite using the same policy.}    
\label{distinct tasks}    
\end{figure*}

\section{Experiments}

In this section, we comprehensively verify the performance of \pGFlowMeta~with three goals in mind: 
1. \pGFlowMeta\ finds higher rewards and more diverse end states faster than state-of-art baselines. 
2. Highlighting that \pGFlowMeta\ can effectively solve the problem of performance degradation when encountering distinct tasks. 
3. Exploring the effect of $\beta$ and $\lambda$.

        

\subsection{Environments Setting}
\label{sec:Environments Setting}



We consider three environments: the environment ``Grid World'' defined in \cite{bengio2021flow} and two Gym Toy Text environments, namely ``Frozen Lake'' and ``Cliff Walking''. We modify the settings of ``Frozen Lake'' and ``Cliff Walking'' to have sparse positive rewards and trajectories forming Directed Acyclic Graphs (DAGs), as shown in Table \ref{tab:summary_of_envs}. Additionally, we introduce two additional goal locations in ``Frozen Lake'' to test the L1 distribution error. Table \ref{tab:summary_of_envs} shows the different physical parameters we set for each environment to induce distinct transitions as different tasks.
For the ``Grid world'' environment, we set different $R_0$ for distinct tasks. For ``Frozen Lake'', we set the different positions of the hole for distinct tasks. For “Cliff Walking”, we set different lengths of cliff walking for distinct tasks. 
Fig. \ref{distinct tasks} provides two examples that illustrate the generation of distinct tasks.

\noindent\paragraph{Frozen Lake} 
The original Frozen Lake environment is represented as a grid where each cell can be one of four possible types: Start (S), Goal (G), Frozen (F), or Hole (H). The agent's objective is to navigate from the start cell to the goal cell while avoiding falling into any of the holes. The agent can only move in four directions: up, down, left, or right, and each move has a chance of slipping and moving the agent in a direction different from the intended direction. The episode ends when the agent reaches the goal or falls into a hole. The rewards are set as follows: reaching a hole (H) or a frozen cell (F) yields a reward of 0, while reaching the goal (G) yields a reward of 1. 

We modify the original Frozen Lake environment to make the rewards positive and the trajectories form DAG. In this modified version, the agent is restricted to taking actions that only move downwards or to the right. The episode ends when the agent reaches the goal, falls into a hole, or reaches the boundary of the lake. We set the rewards for reaching a hole (H) or a frozen cell (F) as $0.1$, and the reward for reaching the goal (G) as $1$.
It should be noted that we create different valid maps to generate distinct tasks, ensuring that each task has a valid path from the start to the goal and defined valid parents for each state.

\noindent\paragraph{Cliff Walking} In the original Cliff Walking environment, an agent is placed in a grid-world environment represented as a grid of cells. The objective of the agent is to navigate from a starting position to a goal position while avoiding falling off a cliff. The board is a $4 \times 12$ matrix, with (using NumPy matrix indexing): $[3, 0]$ as the start at bottom-left, $[3, 11]$ as the goal at bottom-right, and $[3, 1...10]$ as the cliff at bottom-center. The agent can take actions to move in four directions: up, down, left, or right. If the agent steps on the cliff, it returns to the start. An episode terminates when the agent reaches the goal. Each time step incurs a reward of $-1$, and stepping into the cliff incurs a reward of $-100$. Note that there are $3 \times 12 + 2$ possible states in the original environment. In fact, the agent cannot step on the cliff; it can occupy all positions of the first 3 rows, as well as the bottom-left cell and the goal position. 

We modify the original Cliff Walking environment to make the rewards positive and the trajectories form DAG. In the modified environment, the starting position is $[0, 0]$ at the top-left, and the agent is restricted to taking actions that only move downwards or to the right. 
The episode ends when the agent reaches the goal, falls into the cliff, or reaches the boundary of the grid world. We set the rewards to reach the cliff as $0.01$, the goal as $1$, and other positions as $0.1$. We vary the length of the cliff from $5$ to $11$ to generate distinct tasks.

\begin{table*}[ht]
    \centering
    \caption{The parameters to generate different tasks in different environments.}
    \begin{tabular}{cccccccc}
    \hline
        \multirow{2}{*}{Environment} & \multirow{2}{*}{Parameters} & \multicolumn{4}{c}{Goal position} & \multirow{2}{*}{Range} \\
        \cline{3-6}
         &  & Position 1 & Position 2 & Position 3 & Position 4 & \\
    \hline
        Grid world & $R_{0}$ & [0, 0] & [0, 7]& [7, 0] & [7, 7] & [0, 0.1)\\
        Frozen Lake & $Hole$ $position$ & [0, 7] & [7, 4]& [7, 7] & -- & [0, 7] $\times$ [0, 7] \\
        Cliff Walking & $clif\!f$ $length$ & [3, 7] & -- & -- & -- & [8, 11) \\
    \hline    
    \end{tabular}
    \label{tab:summary_of_envs}
\end{table*}
\begin{table*}[ht]
    \centering
    \caption{The hyperparameter settings of \pGFlowMeta\ on diffrent environments.}
    \begin{tabular}{lccc}
    \hline
        \multirow{2}{*}{Hyperparameters} & \multicolumn{2}{c}{Value} \\
        & Grid World & Frozen Lake & Cliff Walking \\
    \hline
        Personalized Learning rate & $1e^{-3}$ & $1e^{-3}$ & $1e^{-3}$ \\
        Global update round $T$ & $30$ & $30$ & $30$\\
        Batch size & $16$ & $16$ &$16$ \\
        Policy network & $[256, 256]$ & $[256, 256]$& $[256, 256]$ \\
        Personalized update round $R$ & $20$ & $20$ & $20$ \\
        Parameter $\lambda$ & $15$ & $15$ & $15$\\
        Learning rate $\eta$ & $0.005$ & $0.005$ & $0.005$ \\
        Discount factor $\gamma$ & $0.99$ & $0.99$ & $0.99$ \\
    \hline
    \end{tabular}
    \label{tab:summary_of_hyperparameters}
\end{table*}

\subsection{Experiments Details}

We conducted our experiments on the NVIDIA Tesla K80 environment. For detailed information on the distinct tasks and parameters used in pGFlowMeta, please refer to Table \ref{tab:summary_of_envs} and Table \ref{tab:summary_of_hyperparameters}. It is important to note that “Batch Size” refers to the size of trajectories rather than individual transitions. Additionally, we provide the key hyperparameter settings for pGFlowMeta in the three environments: “Grid World”, “Frozen Lake”, and “Cliff Walking”. The other parameters of \pGFlowMeta~are basically set to be the same as in \cite{bengio2021flow}. The other compared algorithms were trained for the same number of training steps as \pGFlowMeta. As baselines, we considered the native \textit{GFlowNets} \cite{bengio2021flow}, non meta-RL methods \textit{PPO} \cite{schulman2017proximal}, \textit{\GFlowMeta}, and meta-RL methods like \textit{PEARL} \cite{rakelly2019efficient} to compare their performance \textit{\pGFlowMeta}, which includes both the meta policy (MP) and personalized policy (PP).

 We use the Empirical $L_{1}$ Error \cite{bengio2021flow} to test whether the algorithm generates transitions proportional to the given positive returns. The Empirical $L_{1}$ Error is calculated by Empirical $L_{1}$ Error $ = \mathbb{E}[|p(x)-\pi(x)|]$, where $p(x)=\frac{R(x)}{\sum_{x \in \mathcal{X}} R\left(x\right)}$ represents the known distribution of positive rewards in the environment, and $\pi(x)$ is estimated by repeated sampling and counting frequencies for each possible end state $x$.  Modes Found \cite{bengio2021flow} is used to evaluate the exploration performance. It represents the number of modes with at least one visited state as a function of the total number of visited states. The Averaged Rewards metric is used to measure the performance of an agent in a specific task or environment. It is calculated by taking the mean of the batch end-state rewards from sampled trajectories of test tasks. These evaluation metrics provide valuable insights into the algorithm's performance and its ability to generate appropriate transitions, explore the environment effectively, and achieve high rewards in distinct tasks.

\begin{figure*}[t]
	\centering
	\begin{minipage}{1\linewidth}
		\centering
		\includegraphics[width=0.28\textwidth]{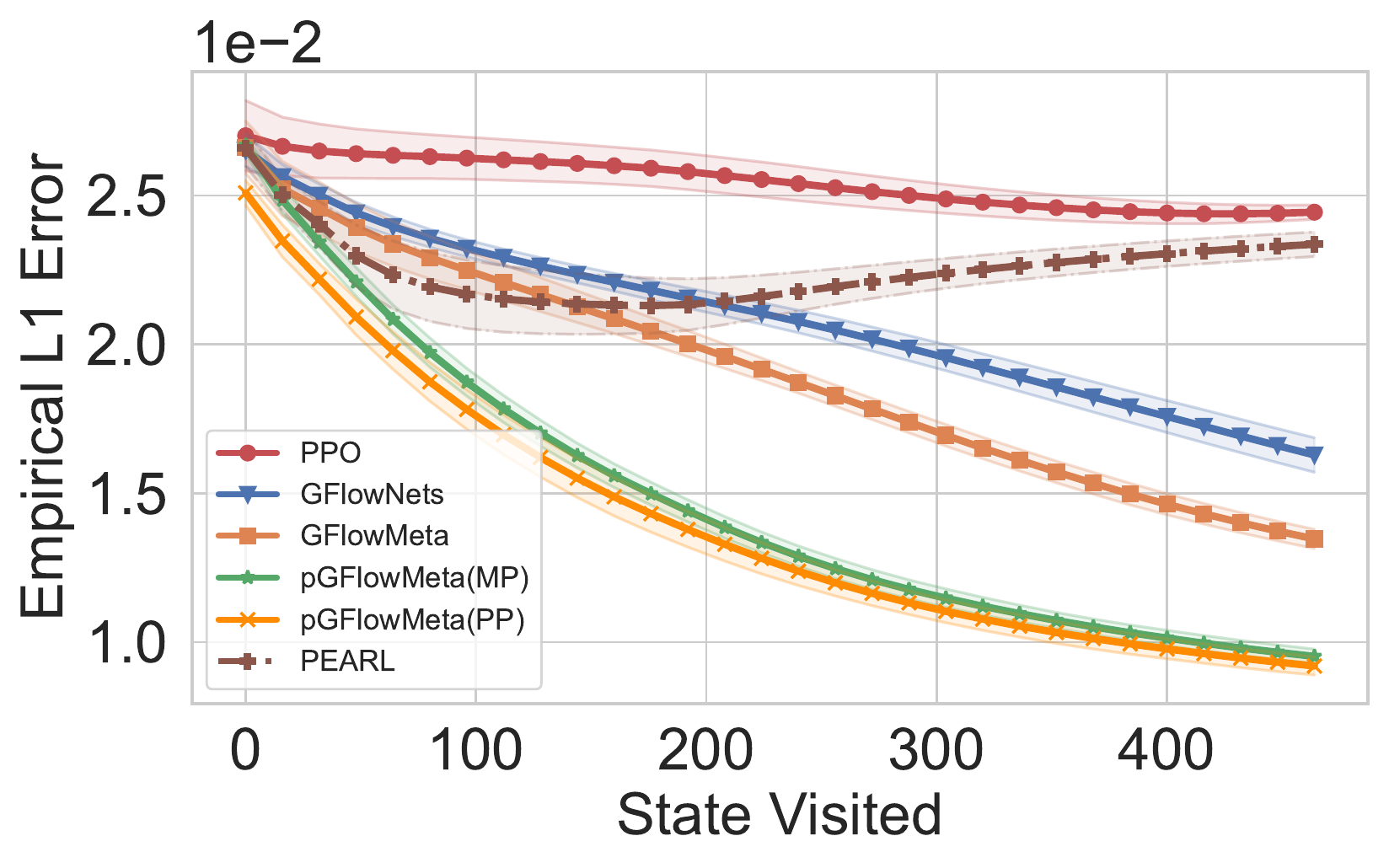}
        \includegraphics[width=0.28\textwidth]{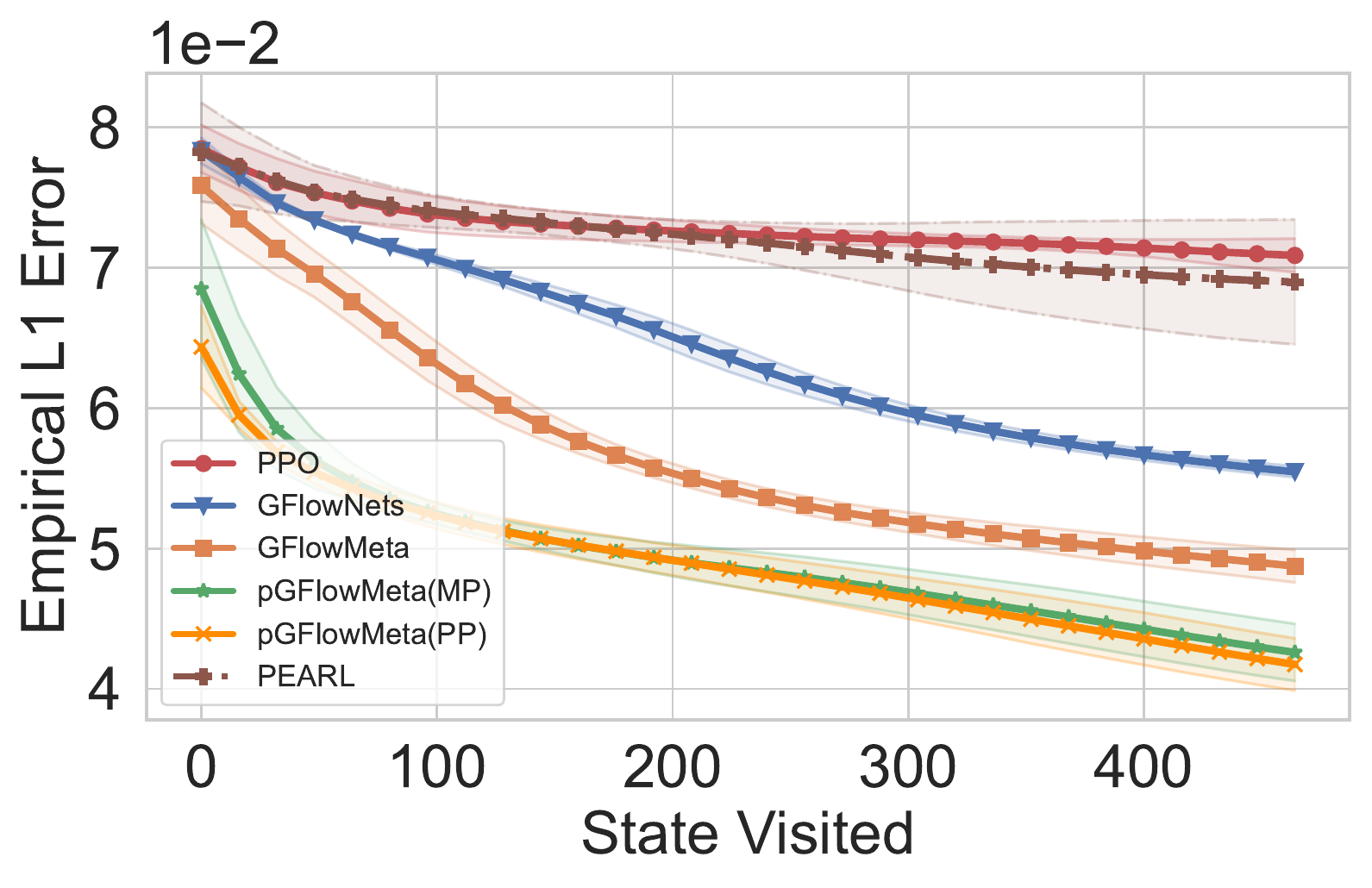}
        \includegraphics[width=0.28\textwidth]{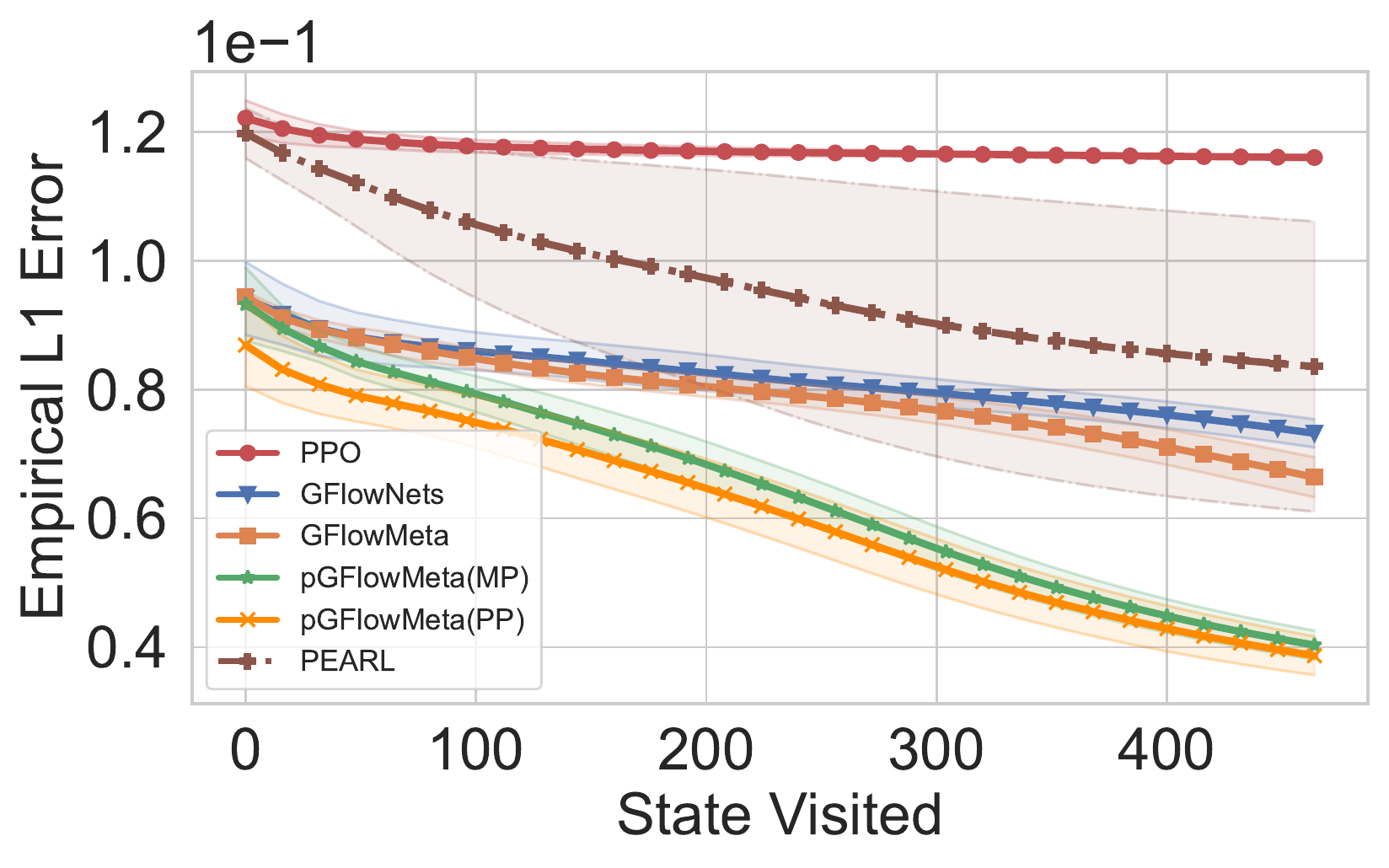}
		\caption{Empirical L1 Error of various algorithms in the environment ‘Grid World’ (left), ‘Frozen Lake’ (middle), ‘Cliff Walking’ (right).}
		\label{fig:comparation_envs1}
	\end{minipage}
 
	\begin{minipage}{1\linewidth}
		\centering
		\includegraphics[width=0.28\textwidth]{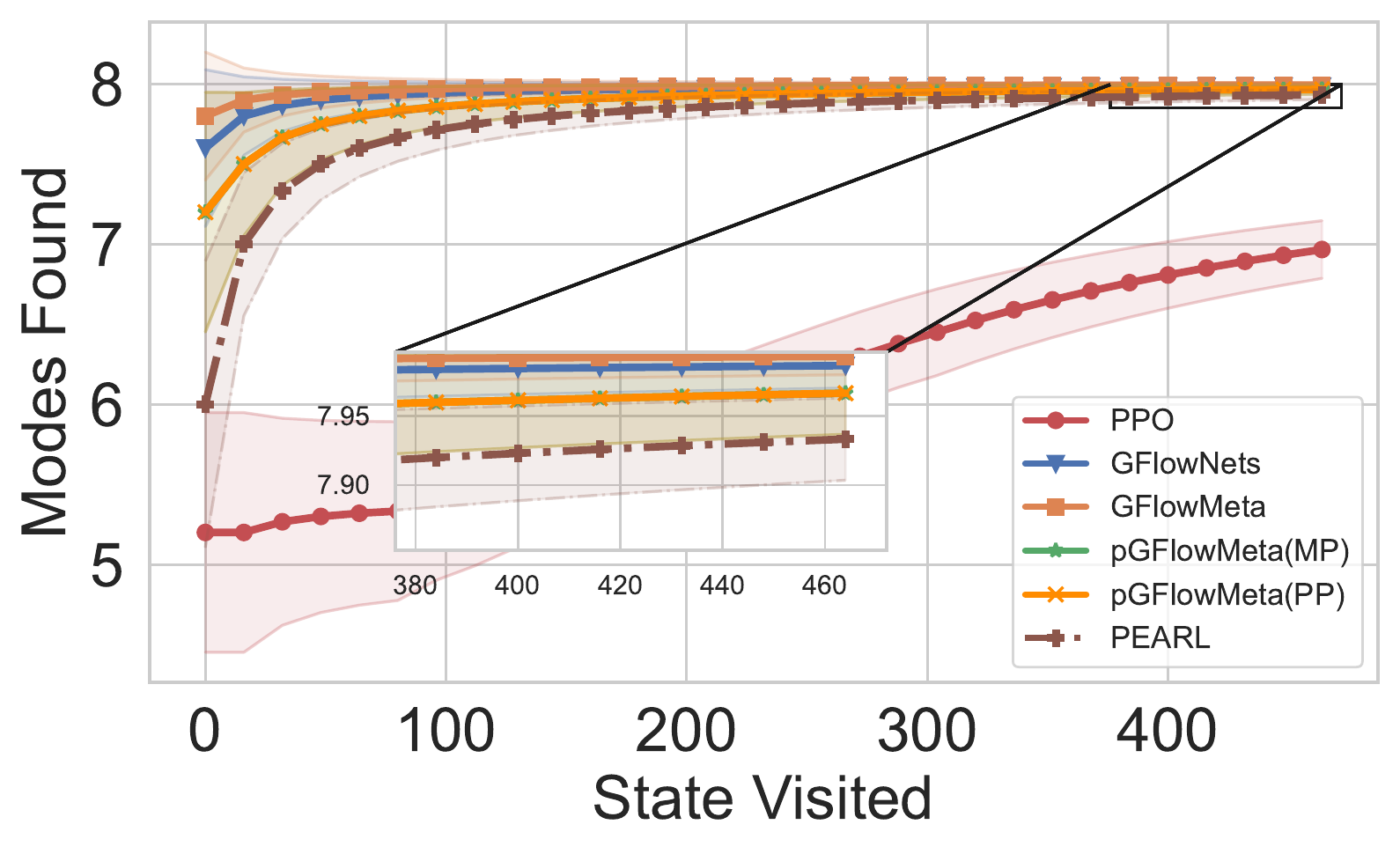}
    \includegraphics[width=0.28\textwidth]{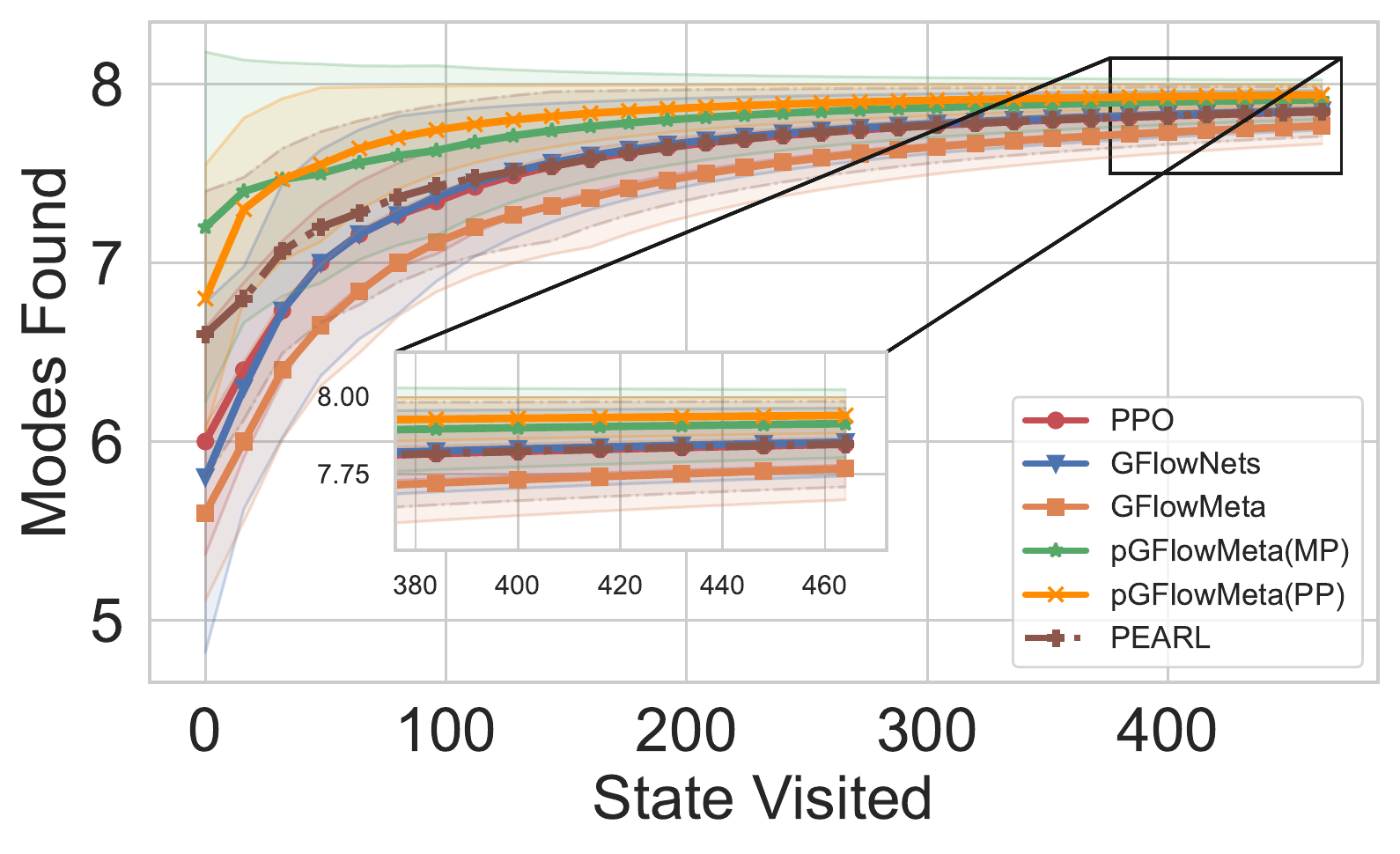}
    \includegraphics[width=0.28\textwidth]{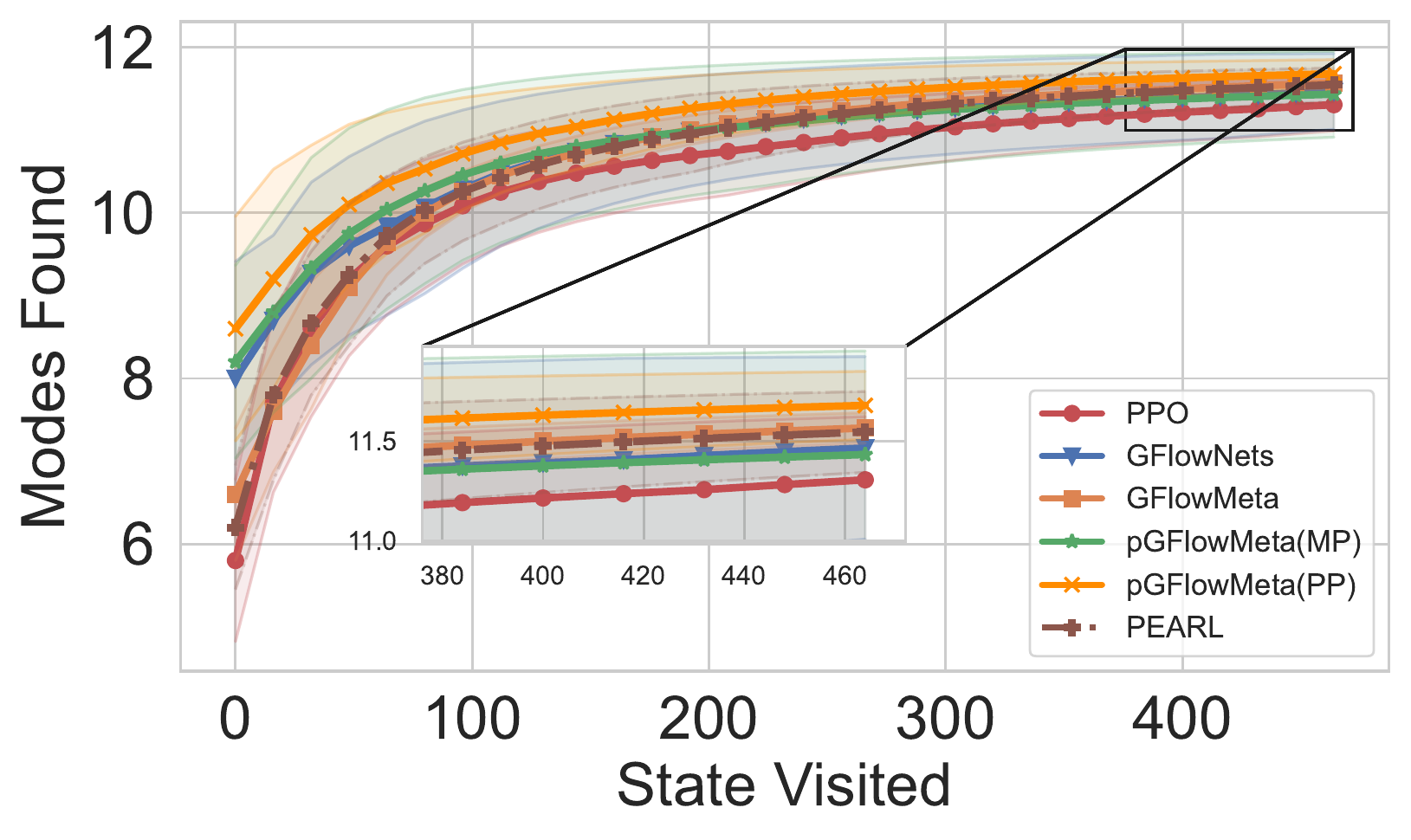}
    \caption{Modes Found of various algorithms in the environment 'Grid World' (left), 'Frozen Lake' (middle), 'Cliff Walking' (right).}
    \label{fig:comparation_envs2}
	\end{minipage}
	\begin{minipage}{1\linewidth}
		\centering
		\includegraphics[width=0.28\textwidth]{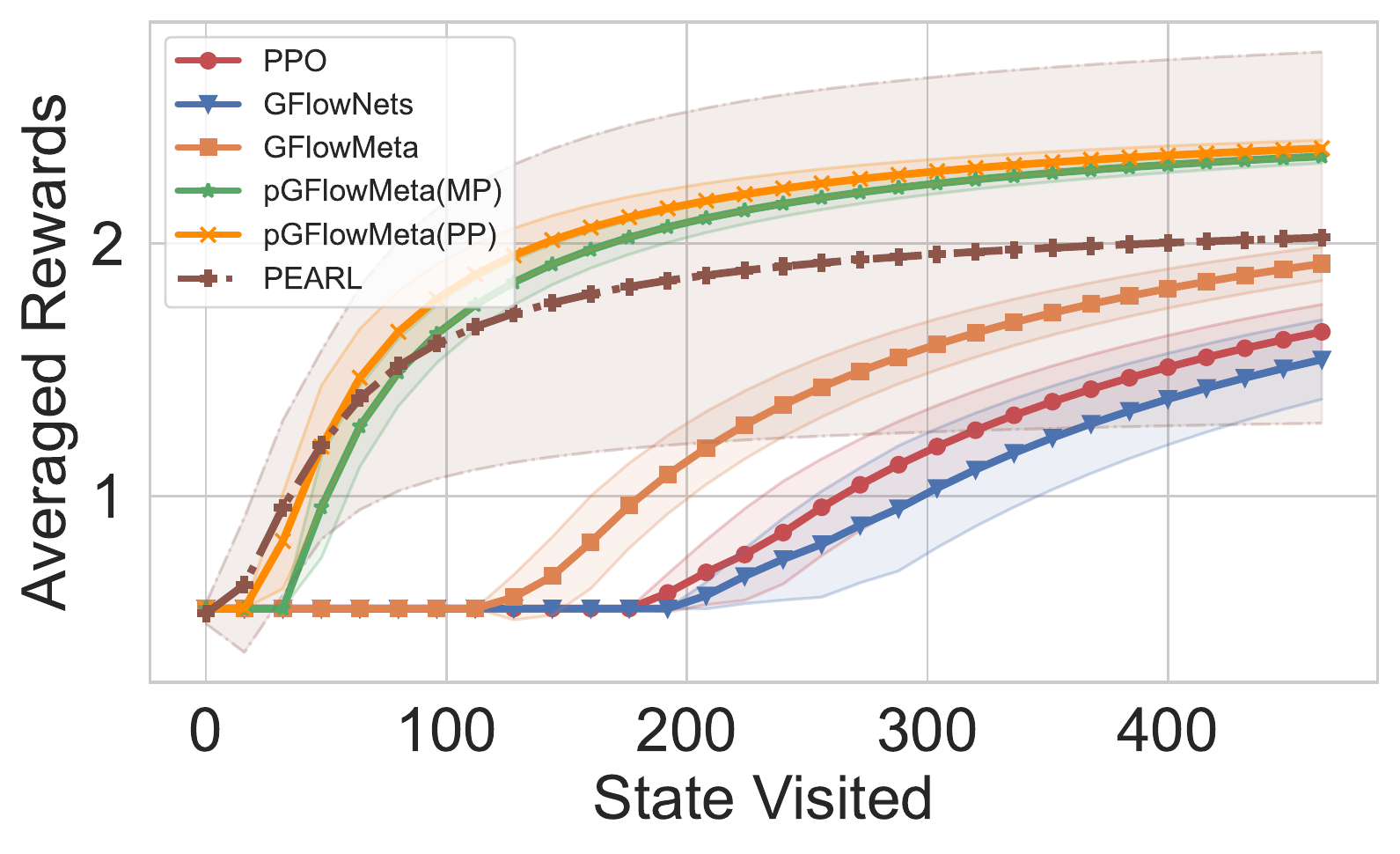}
    \includegraphics[width=0.3\textwidth]{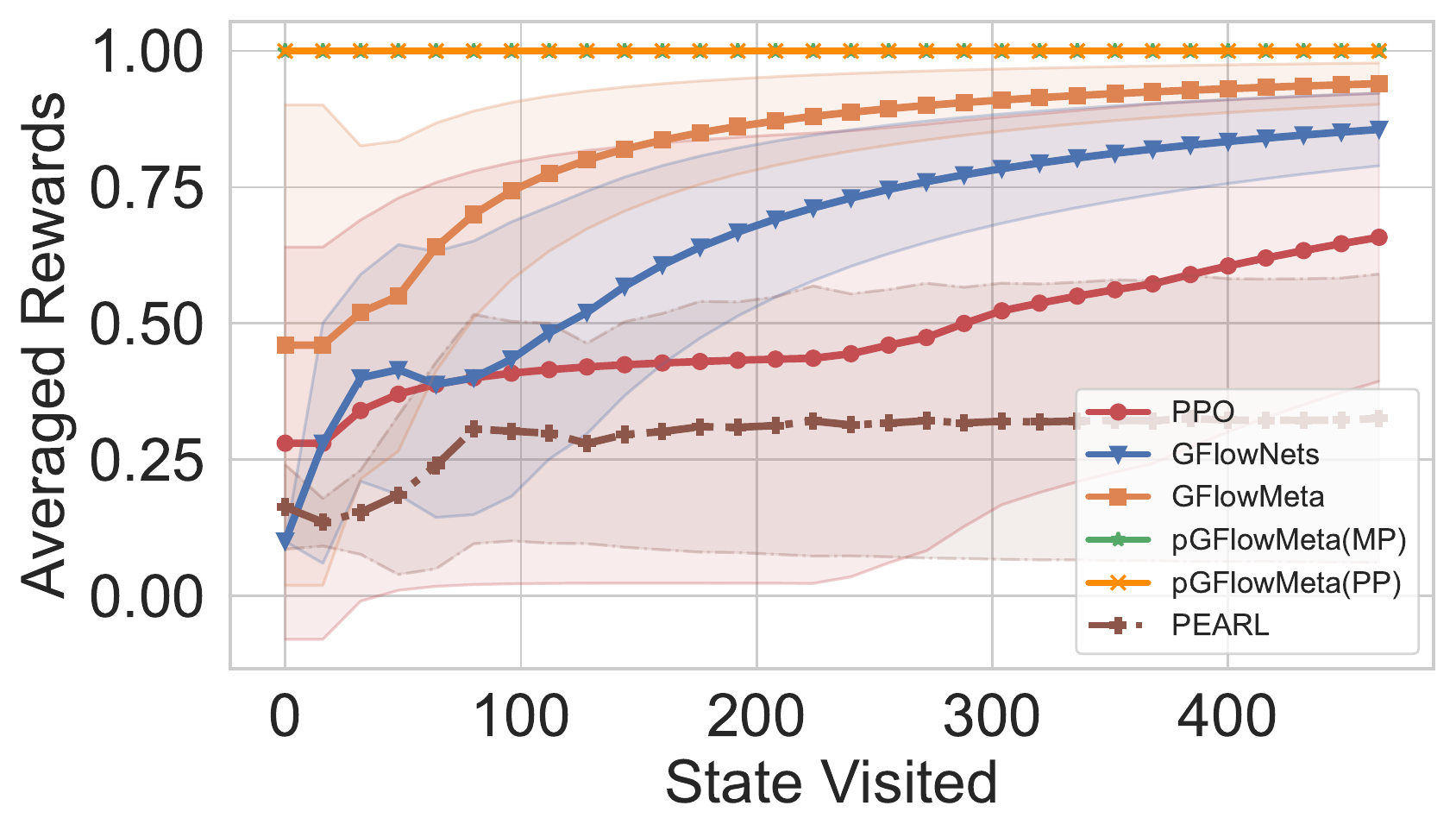}
    \includegraphics[width=0.28\textwidth]{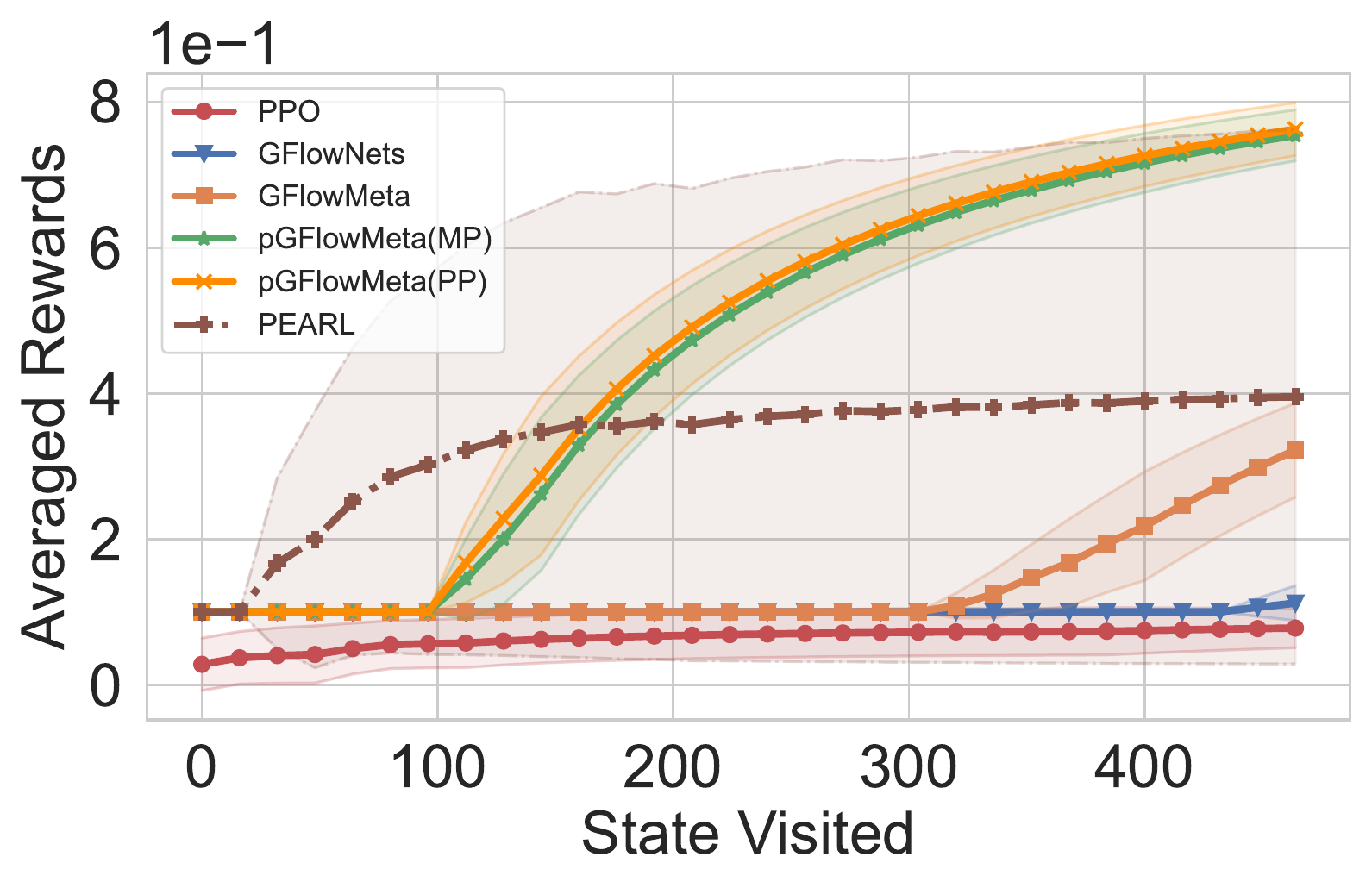}
    \caption{Averaged rewards of various algorithms in the environment ‘Grid World’ (left), ‘Frozen Lake’ (middle), ‘Cliff Walking’ (right).}
    \label{fig:comparation_envs3}
	\end{minipage}
\end{figure*}

\subsection{Comparison to the State-Of-The-Art Methods}

In this section, we compare the performance of state-of-the-art methods with the proposed \pGFlowMeta\ methods on three environments. 
We have described these environments in detail in the \ref{sec:Environments Setting}. 

\paragraph{Empirical $L_1$ Error} Fig. \ref{fig:comparation_envs1} shows the evaluation results of Empirical $L_{1}$ Error on three environments. The figure demonstrates that GFlowNets, \GFlowMeta, \pGFlowMeta\ achieve lower Empirical $L_{1}$ Error compared to PEARL and PPO methods. This suggests that the methods within the GFlowNets class have the capability to generate transitions proportional to positive rewards, which is valuable in exploration tasks. Traditional RL methods typically generate transitions associated with the highest reward, while these methods can generate transitions proportional to positive rewards.
Among the methods compared, \pGFlowMeta\ shows the lowest Empirical $L_{1}$ Error, which aligns with the experimental expectations. It shows that \pGFlowMeta\ effectively adapts to distinct tasks compared to GFlowMeta.

\paragraph{Modes Found} 
Fig. \ref{fig:comparation_envs2} illustrates the Modes Found in three different environments. It depicts the number of modes visited at least once for various methods, as the number of visited states increases. It is important to note that the total number of Modes Found is primarily influenced by the specific design of the reward function. If there are $n$ end states with positive rewards, the total number of Modes Found equals $n$. From Fig. \ref{fig:comparation_envs2}, we can find that: (1) \pGFlowMeta\ can almost achieve faster visits to all the end states in different environments. (2) PPO almost achieved the slowest visits to all the end states in different environments. That is, \pGFlowMeta\ has the best exploration in general situations, and the PPO algorithm usually shows less exploration than other methods.

\paragraph{Rewards} In addition to exploration and sampling transitions proportional to positive rewards, the average rewards of different methods are shown in Fig. \ref{fig:comparation_envs3}.
The figure demonstrates that \pGFlowMeta\ achieves the best average rewards compared to other methods. While the primary objective of \pGFlowMeta\ is to sample transitions proportional to positive rewards, it can also be configured to prioritize achieving goals with the highest rewards.  It is important to note that this prioritization differs from the exploration setting. In contrast, \pGFlowMeta\ has the capability to sample transitions using a deterministic policy when applied to test tasks.



\subsection{Effect of $\beta$ and $\lambda$ values}

\begin{figure*}[t]
    \centering
    \subfigure{
        \centering
        \includegraphics[width=0.23\textwidth]{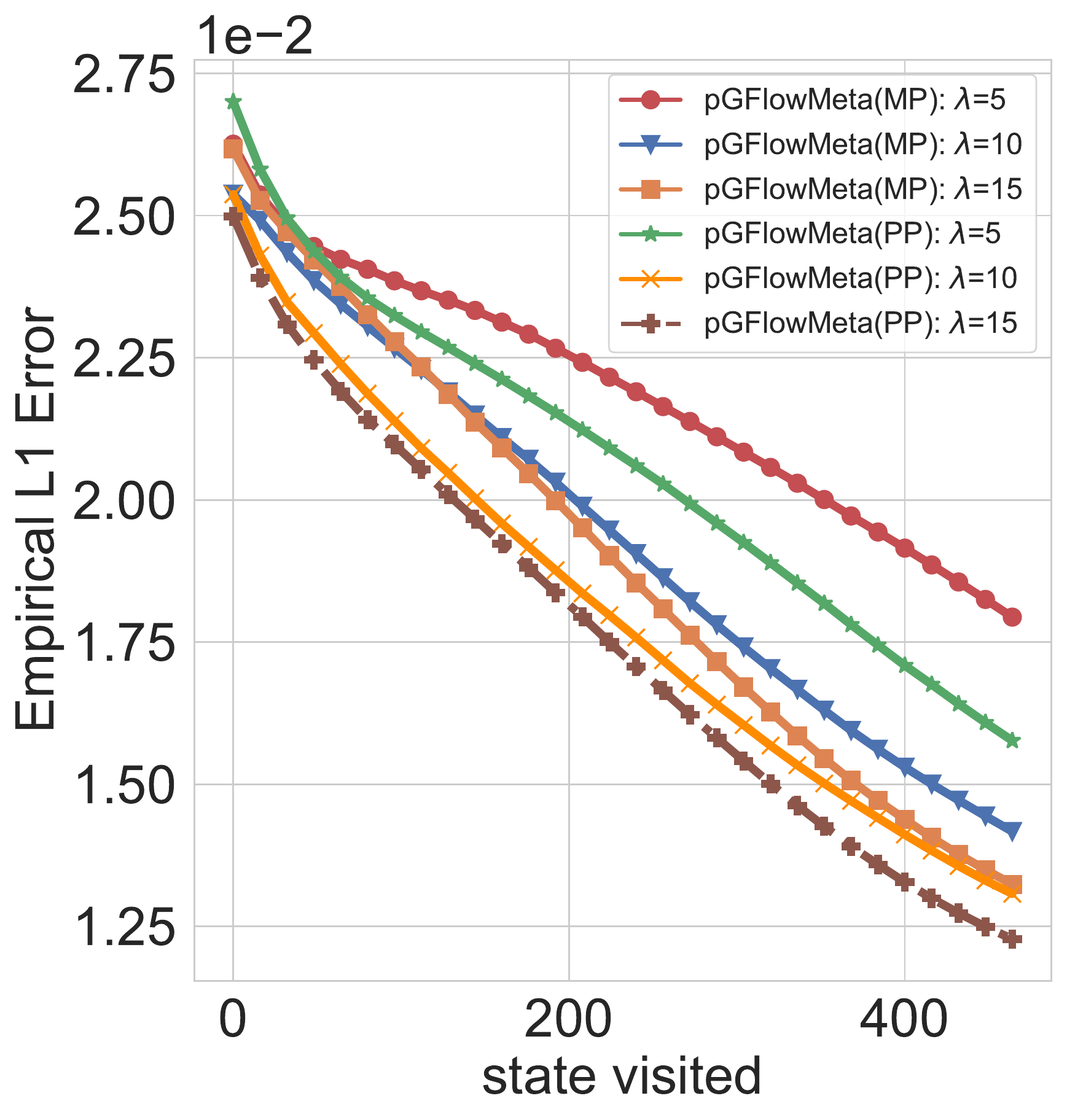}
        \label{fig:subfiga1}}     
    \subfigure{
        \centering \includegraphics[width=0.224\textwidth]{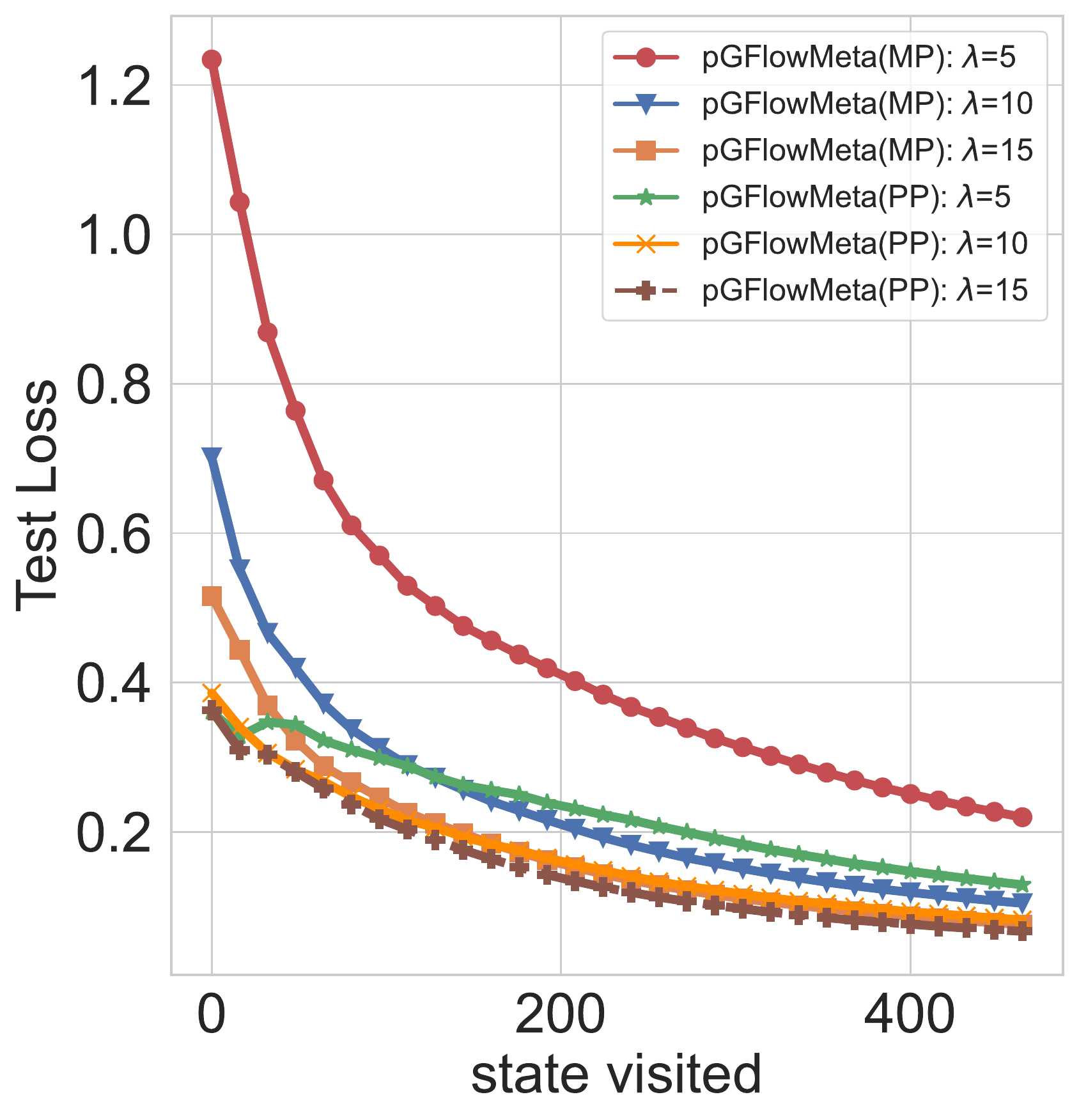}
    }
    \subfigure{
        \centering 
    \includegraphics[width=0.23\textwidth]{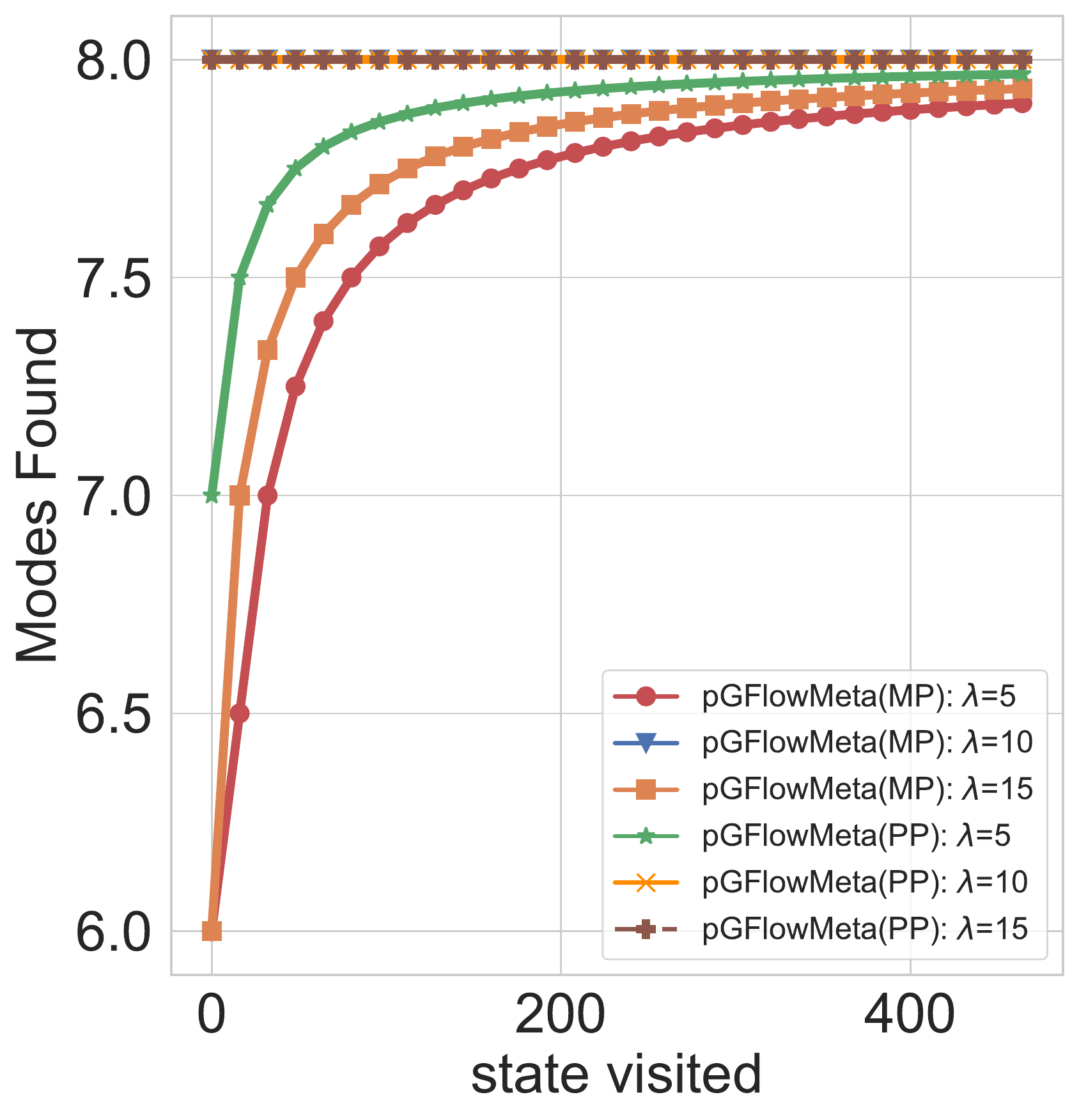}
    }
    \subfigure{
        \centering \includegraphics[width=0.224\textwidth]{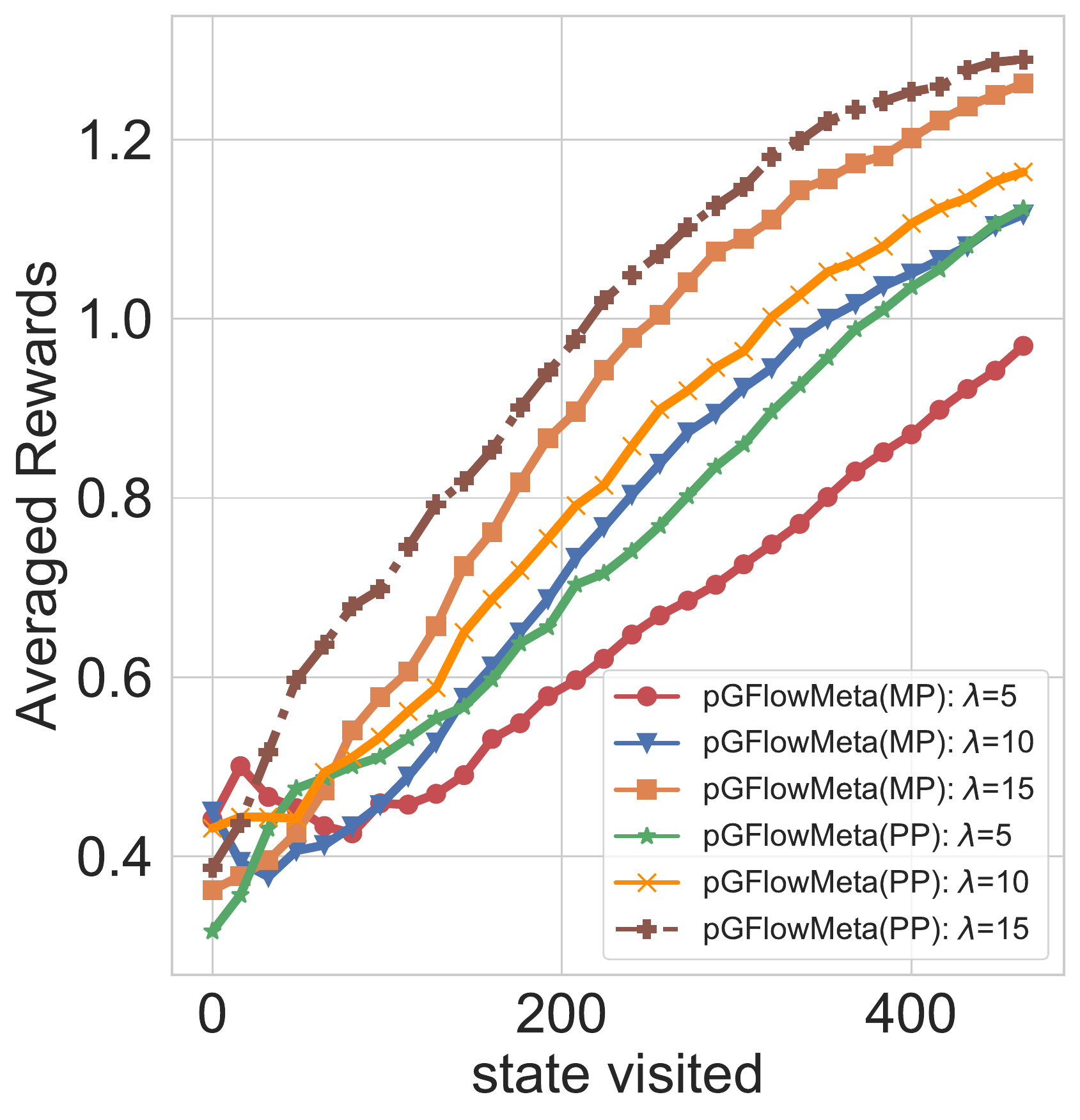}
    }
    \caption{Effect of $\lambda$ on the convergence of \pGFlowMeta\ in ‘Grid World’ environment. $R$ = 20, $\beta$ = 1. }
    \label{fig:lambda}
\end{figure*}

Another important aspect that requires investigation is the impact of the $\beta$ and $\lambda$ values on \pGFlowMeta. We run these experiments in the ‘Grid World’ environment. In Fig. \ref{fig:lambda},  observe the performance of \pGFlowMeta\ with different values of $\lambda$ while keeping $B = 16$, $R = 20$, $\eta = 0.005$, $\beta = 1$. The graph indicates that a larger $\lambda$ value ($\lambda = 15$) leads to faster convergence for \pGFlowMeta. Therefore, we select $\lambda$ = 15 for this paper. However, it is worth noting that larger $\lambda$ can also cause \pGFlowMeta\ to diverge. Hence, it is essential to carefully select $\lambda$ for different environments. 

\begin{figure*}[t]
    \centering
    \subfigure{
        \centering
     \includegraphics[width=0.23\textwidth]{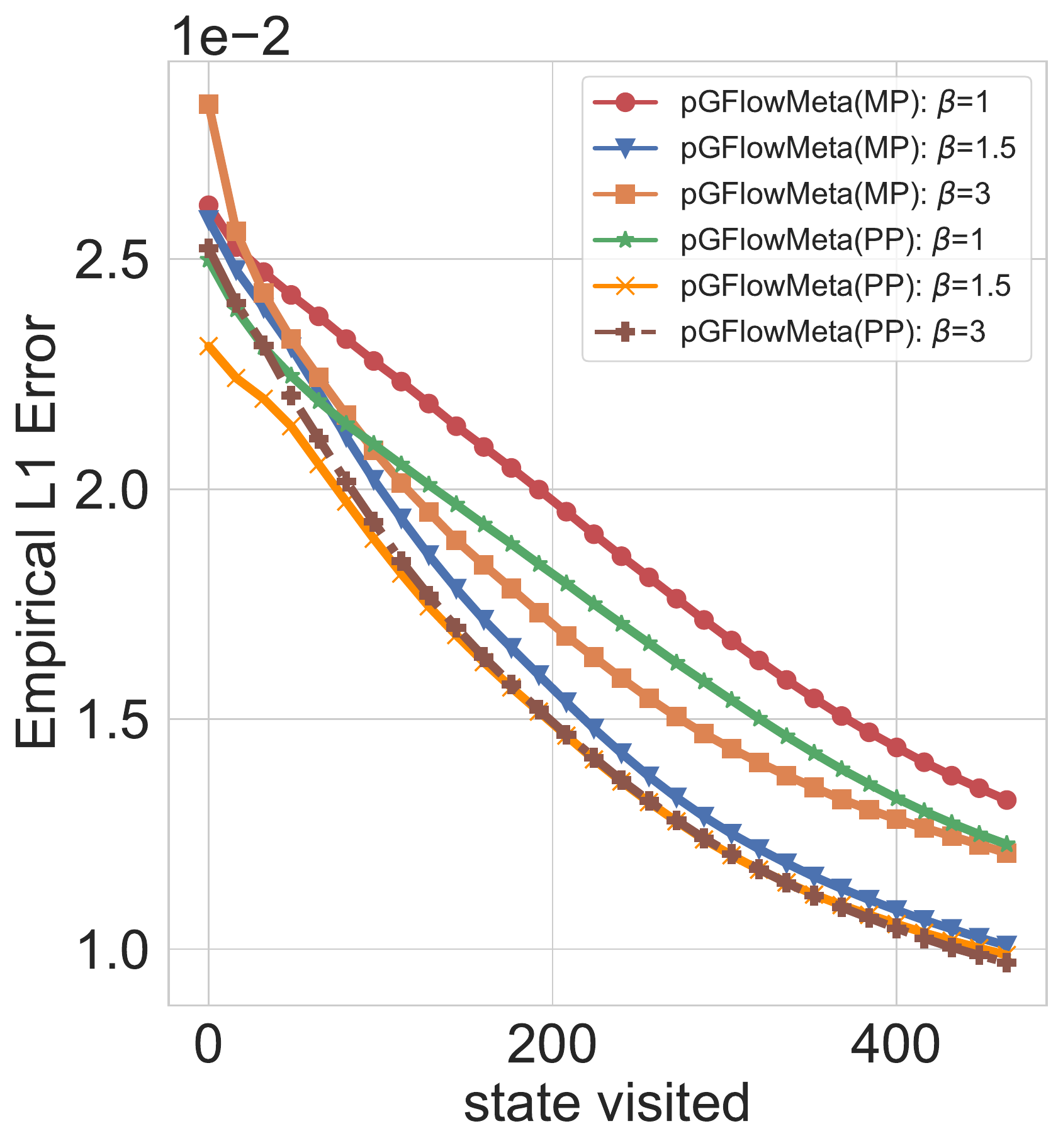}
     }     
    \subfigure{
        \centering \includegraphics[width=0.224\textwidth]{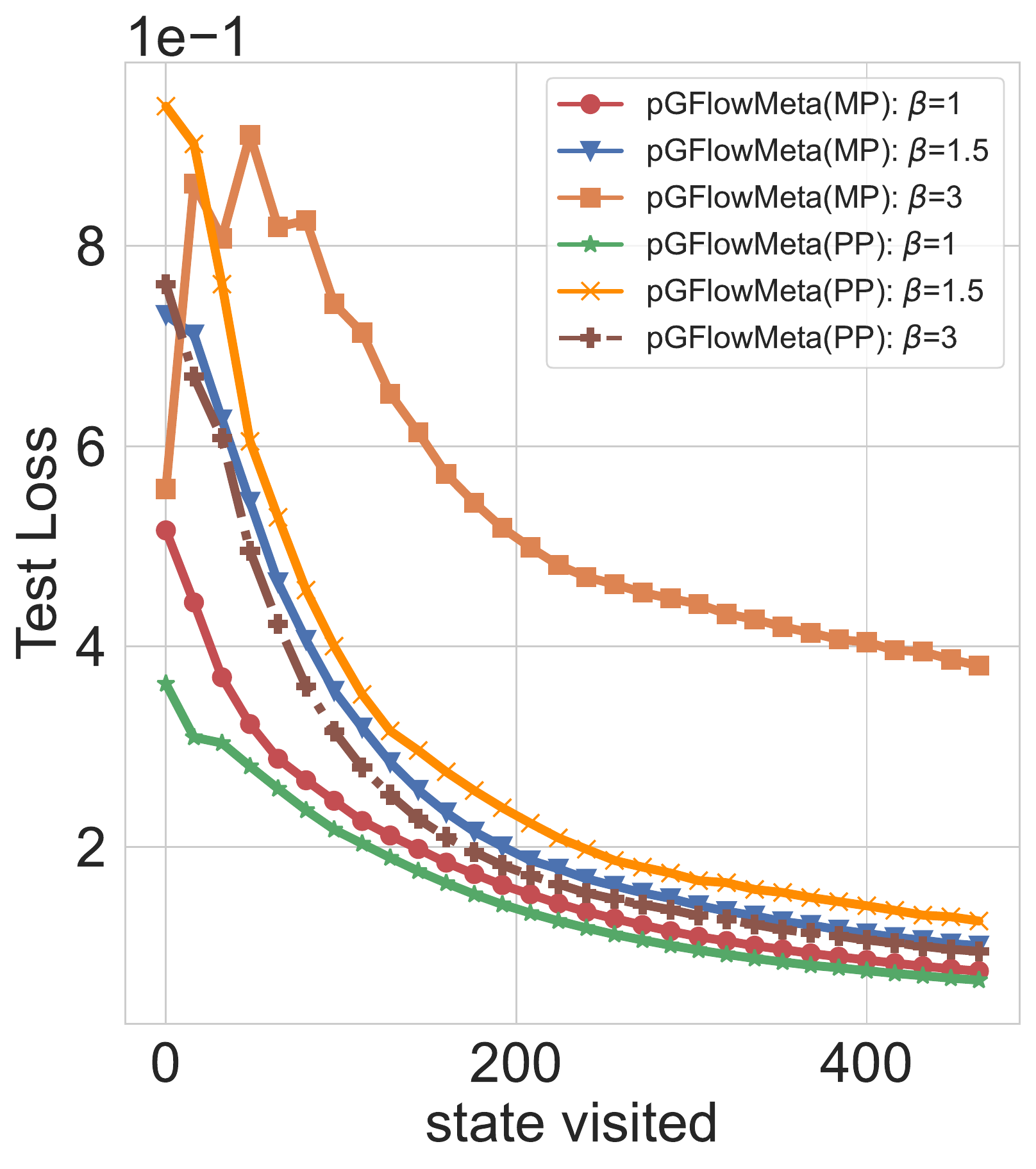}
    }
    \subfigure{
        \centering 
    \includegraphics[width=0.23\textwidth]{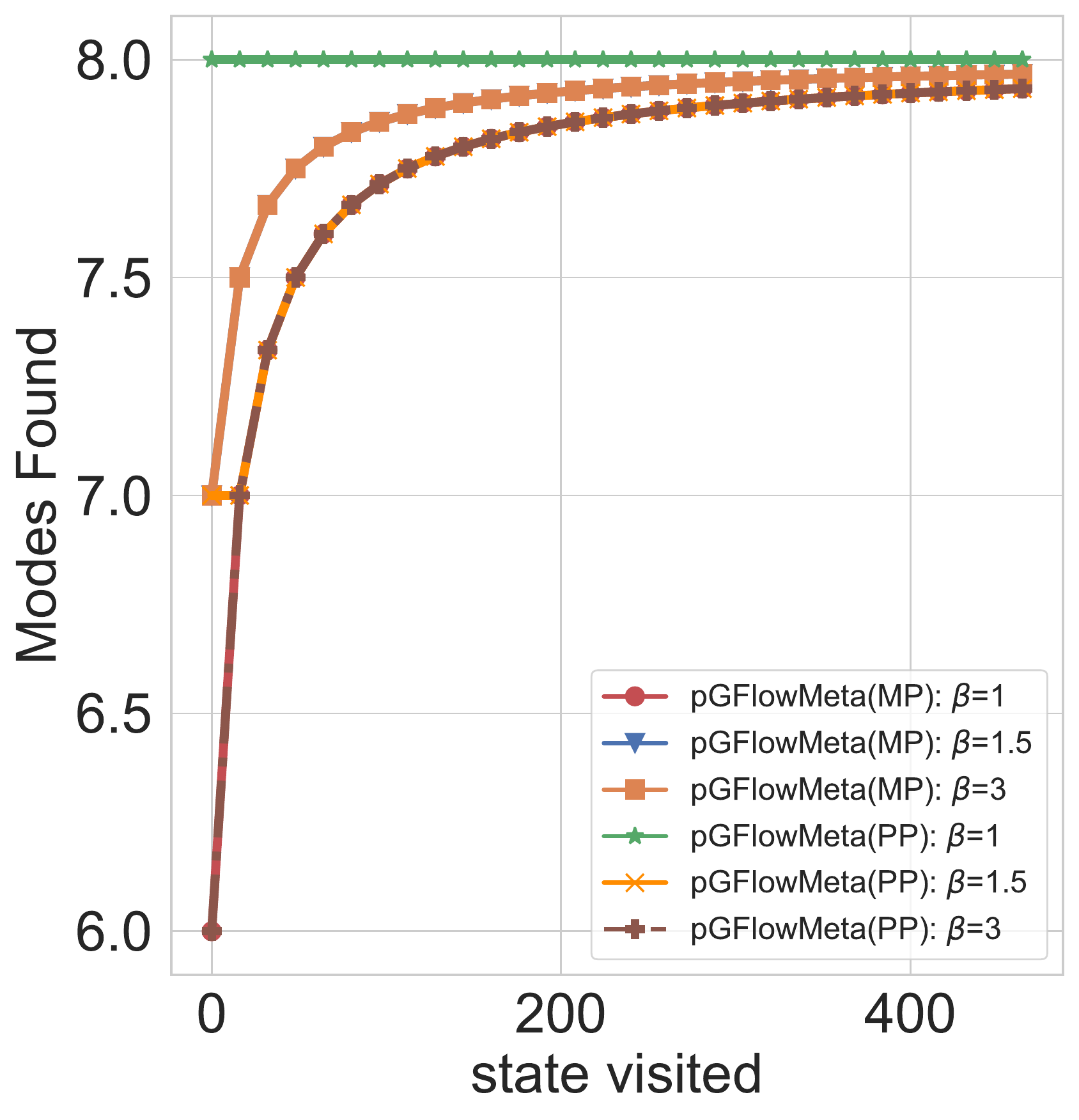}
    }
    \subfigure{
        \centering \includegraphics[width=0.224\textwidth]{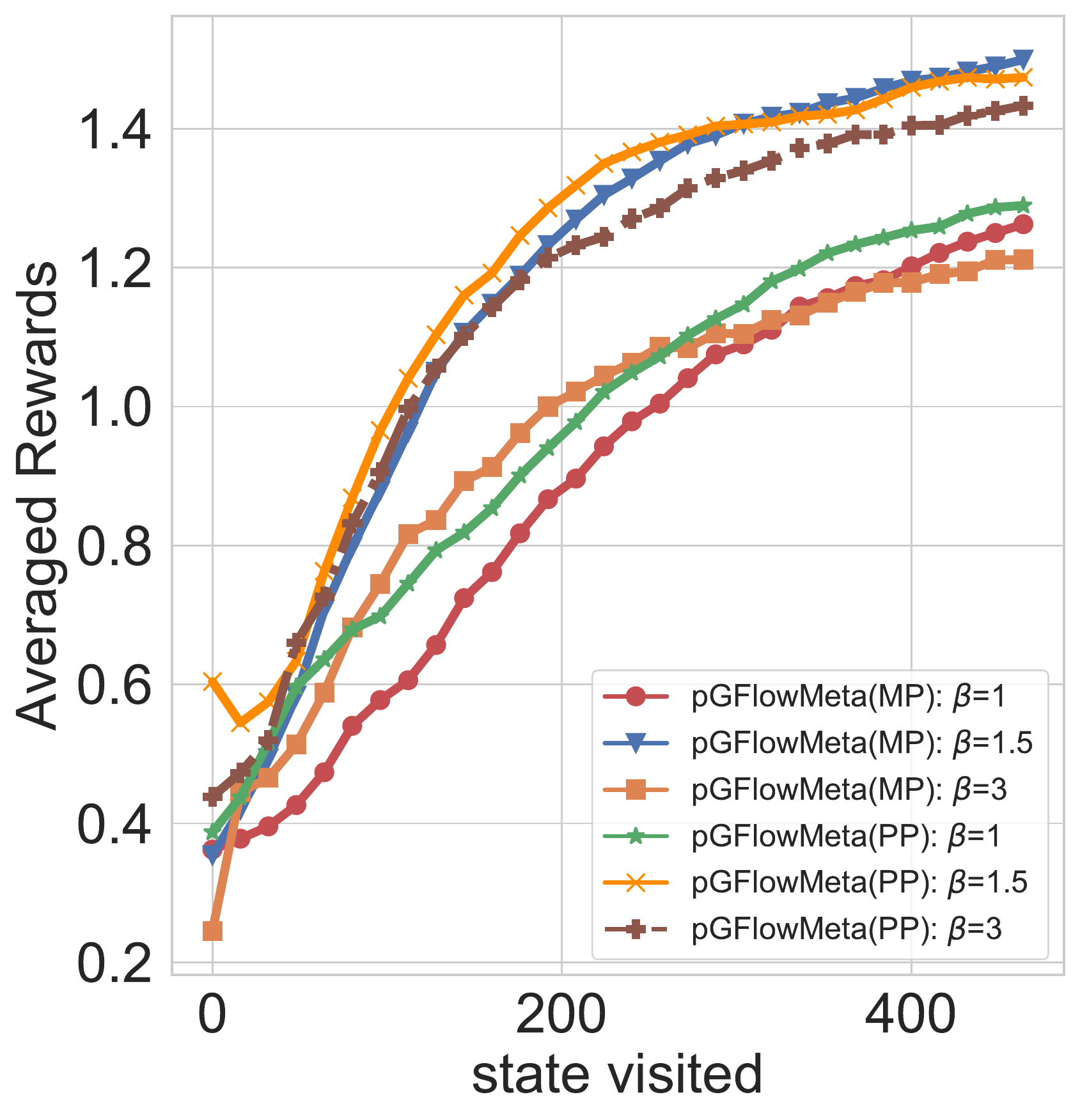}
    }
    \caption{Effect of $\beta$ on the convergence of \pGFlowMeta\ in ‘Grid World’ environment. $R$ = 20, $\lambda$ = 15.}
    \label{fig:beta}
\end{figure*}

In Fig. \ref{fig:beta}, we explore the influence of $\beta$ ($\beta \geq 1$) on both the personalized and meta policies in \pGFlowMeta, using $B = 16$, $R = 20$, $\eta = 0.005$, $\lambda = 15$. Here, $\beta = 1$ implies that the aggregation of the meta policy in \pGFlowMeta\ is the same as that in \GFlowMeta. As depicted in the figure, different evaluation indices exhibit distinct optimal $\beta$ values. Notably, a $\beta$ value of 3 yields the best performance in terms of Empirical L1 Error, while $\beta = 1$ demonstrates the optimal performance for test loss. Hence, the selection of $\beta$ should be done thoughtfully, taking into account the specific use case or objective at hand. Furthermore, it is important to highlight that the personalized policy consistently outperforms the meta policy in \pGFlowMeta, regardless of the chosen metric. This can be attributed to the fact that the personalized policy is capable of adapting to the unique characteristics and requirements of individual tasks, whereas the meta policy serves as a general policy applied to all tasks.

\subsection{Ablation Experiments}
We conducted a comparison of the performance of \pGFlowMeta\ with GFlowNets, GFlowNets$\ast$, and GFlowMeta, and the results are presented in Table \ref{tab:ablation}, which shows the averaged rewards achieved by each algorithm. GFlowNets$\ast$ represents the averaged rewards obtained by the optimal GFlowNets policy for each task, while GflowNets represents the averaged rewards achieved by the GFlowNets policy across multiple tasks. 

From the table, we can observe that our personalization policy (PP) is closer to the optimal policy, indicating that it performs well in adapting to the specific requirements of individual tasks. Additionally, our meta policy (MP) outperforms the other algorithms, suggesting that it exhibits superior overall performance. Fig. \ref{fig:comparation_envs1}, \ref{fig:comparation_envs2}, \ref{fig:comparation_envs3}  provides further insights into the comparison, showing that the personalized policy is more suitable for its specific task than the meta policy. This highlights the generalization ability of the personalized policy within the constraints imposed by the meta policy.  Furthermore, the performance of meta policy in \pGFlowMeta\ is better than that in \GFlowMeta, indicating that personalized policy can improve the performance of meta policy in \pGFlowMeta.


\subsection{Performance on Similar / Distinct Tasks }

In this experiment, we conducted a comparison between \pGFlowMeta\ and GflowMeta methods on two types of tasks: distinct tasks and similar tasks. The purpose was to assess the performance difference between GFlowMeta and pGFlowMeta in these task scenarios. The experiments were repeated five times,  and the results were averaged. For each run,  we generated distinct tasks according to the instructions provided in Section \ref{sec:Environments Setting} and Table \ref{tab:summary_of_envs}. To be fair, we used the different parameters of tasks to create similar tasks for each experiment run. 

\begin{table}[!t]
    \centering
    \caption{The averaged reward comparison results for each algorithm.}
    \scalebox{0.8}{
    \begin{tabular}{lccc}
    \hline
    \specialrule{0em}{1pt}{1pt}
        Algorithms & Grid World & Frozen Lake & Cliff Walking \\
    \specialrule{0em}{1pt}{1pt}
    \hline
        GFlowNets$\ast$ & 2.42 ± 0.03 & 1.00 $\pm$ 0.00 & 0.77 $\pm$ 0.04 \\
        GFlowNets & 1.42 $\pm$ 0.07 & 0.86 $\pm$ 0.07 & 0.11 $\pm$ 0.02 \\ 
        GFlowMeta & 1.92 $\pm$ 0.07 & 0.94 $\pm$ 0.04 &         0.32 $\pm$ 0.06 \\
        \pGFlowMeta\ (MP) & 2.34 $\pm$ 0.03 & 1.00 $\pm$ 0.00 & 
        0.75 $\pm$ 0.03 \\
        \pGFlowMeta\ (PP) & \textbf{2.38 $\pm$ 0.03} & \textbf{1.00 $\pm$ 0.00} & \textbf{0.76 $\pm$ 0.04} \\
    \hline
    \end{tabular}
    }
    \label{tab:ablation}
\end{table}
\begin{table}
    \centering
    \caption{The averaged reward comparison results on similar and distinct tasks for each algorithm.}
    \scalebox{0.78}{
    \begin{tabular}{llcc}
    \hline
    \specialrule{0em}{1pt}{1pt}
        Environments & Algorithms & Similar Tasks & Distinct Tasks \\
    \specialrule{0em}{1pt}{1pt}
    \hline
        \multirow{3}{*}{Grid World} & GFlowMeta & 1.96 $\pm$ 0.08 & 1.92 $\pm$ 0.07\\
        & \pGFlowMeta\ (MP) & 2.36 $\pm$ 0.03 & 2.34 $\pm$ 0.03 \\
        & \pGFlowMeta\ (PP) & \textbf{2.38 $\pm$ 0.04} & \textbf{2.38 $\pm$ 0.03}\\
    \hline
        \multirow{3}{*}{Frozen Lake} & GFlowMeta & 0.96 $\pm$ 0.05 & 0.94 $\pm$ 0.04\\
        & \pGFlowMeta\ (MP) & \textbf{1.00 $\pm$ 0.00} & 1.00 $\pm$ 0.00\\
        & \pGFlowMeta\ (PP) & 0.99(8)  $\pm$ 0.00(2) & \textbf{1.00 $\pm$ 0.00} \\
    \hline
        \multirow{3}{*}{Cliff Walking} & GFlowMeta & 0.36 $\pm$ 0.13 & 0.32 $\pm$ 0.06 \\
        & \pGFlowMeta\ (MP) & \textbf{0.78 $\pm$ 0.01} & 0.75 $\pm$ 0.03\\
        & \pGFlowMeta\ (PP) & 0.76 $\pm$0.04 & \textbf{0.76 $\pm$ 0.04}\\
    \hline
    \end{tabular}
    }
    \label{tab:ablation2}
\end{table}
\begin{figure}[!t]
\centering
\includegraphics[width=1\linewidth]{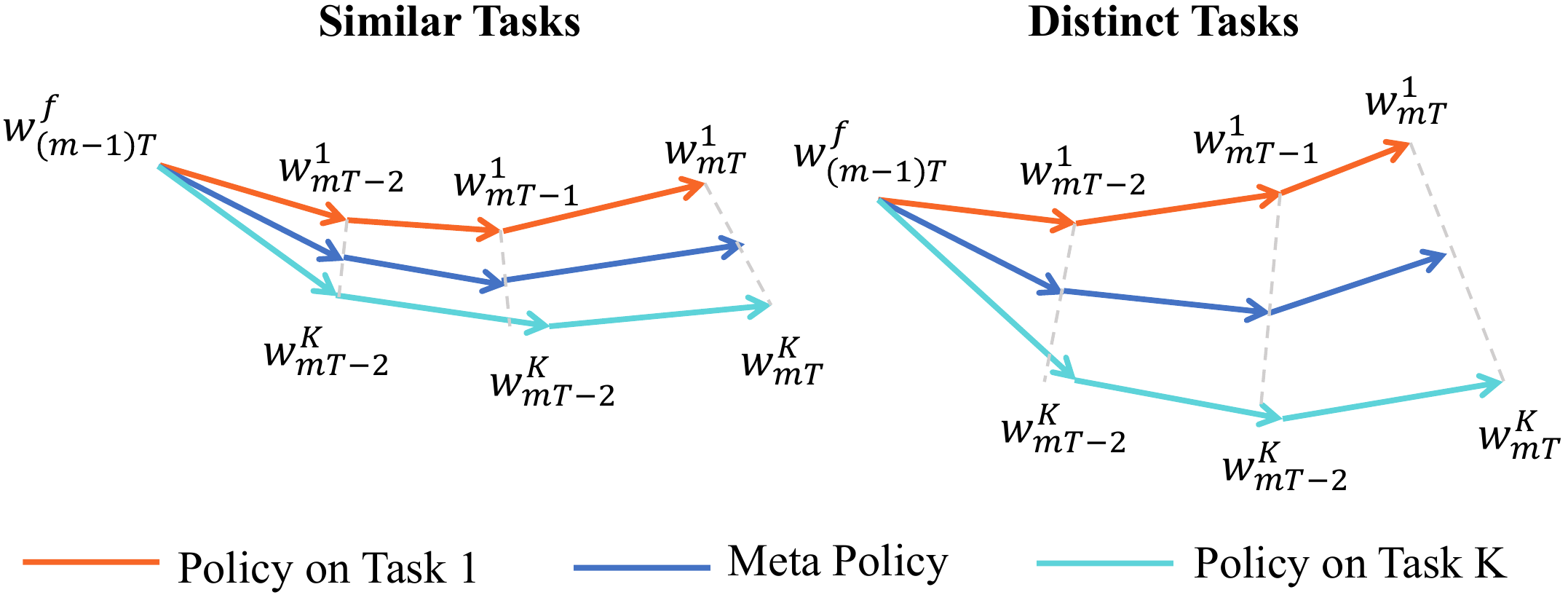}
\caption{Illustration of the weight divergence for the policy with similar tasks and distinct tasks.}
\label{fig:intro}
\end{figure}

Table \ref{tab:ablation2} presents the averaged rewards for GFlowMeta and \pGFlowMeta, with the best results on each environment highlighted in bold. As expected, both \pGFlowMeta\ and \GFlowMeta\ experienced a decrease in performance when dealing with distinct tasks. This decrease can be attributed to the differences between the task distributions. Fig. \ref{fig:intro} provides an illustration to help understand this phenomenon. When tasks are similar, the divergence between the Policy on Task $k$ and Meta Policy is small. Even after the $m$-th synchronization, the Policy on Task $k$ remains still close to Meta Policy. However, when tasks are distinct, the divergence between Policy on Task $k$ and Meta Policy becomes much larger and accumulates rapidly due to the differences in data distribution. This results in a significant increase in divergence. Furthermore, \pGFlowMeta\ demonstrates lower performance degradation compared to GflowMeta, indicating its higher robustness to distinct tasks. This showcases the advantage of \pGFlowMeta\ in adapting and maintaining performance even when faced with varying and diverse task scenarios.



\section{Conclusion}
In this paper, we propose a novel approach called \pGFlowMeta, which is a personalized Meta Generative Flow Network designed to achieve personalization on distinct tasks. The objective of \pGFlowMeta\ is to learn both a meta policy and personalized policies for all tasks and specific tasks, respectively. To achieve this, we incorporate a personalization constraint into the objective function, encouraging the development of personalized policies within the framework of the meta policy. The primary aim of \pGFlowMeta\ is to generate samples proportional to specific positive rewards, ensuring that each task's policy is tailored to its unique requirements. An alternating minimization method is proposed to separate the learning processes of personalized and meta policies. Additionally, theoretical analysis demonstrates that our algorithm converges sublinearly with the iteration number and provides an upper bound on the difference between the personalized policies and the meta policy. Experimental results showcase the effectiveness of \pGFlowMeta, as it outperforms several advanced algorithms in discrete control tasks. These results highlight the capability of \pGFlowMeta\ to achieve superior performance and demonstrate the effectiveness of personalizing policies for individual tasks within the meta policy framework.






\appendix
\section*{Declaration of competing interest}
The authors declare that they have no known competing financial interests or personal relationships that could have appeared to influence the work reported in this paper.
\section*{Data availability}
Data will be made available on request.

\section*{Acknowledgments}
This work is supported by the National Natural Science Foundation of China under No. 12288201, China National Postdoctoral Program for Innovative Talents under No. BX2021346 and China Postdoctoral Science Foundation under No. 2022M713316.

\printcredits

\section{Proof of Main Theorems}
{
\subsection{Some Useful Results}
\begin{fact}[Jensen's inequality]\label{jensen}
 For arbitrary vector $x_i \in \mathbb{R}^d, i=1,\cdots, M$, the following inequalities hold
\begin{equation}
\begin{aligned}
    & \left\|\frac{1}{M}\sum_{i=1}^M x_i\right\|^2 \leq  \frac{1}{M}\sum_{i=1}^M\left\|x_i\right\|^2\\
     & \left\|\sum_{i=1}^M x_i\right\|^2 \leq  M\sum_{i=1}^M\left\|x_i\right\|^2.\\
\end{aligned}
\end{equation}
\end{fact}

\begin{fact}[Young's inequality]\label{triangle}
 For any vector $x_1, x_2 \in \mathbb{R}^d$ and any $\varepsilon>0$, we have
\begin{equation}
\begin{aligned}
    2\ip{x_1}{x_2} &\leq \frac{1}{\varepsilon}\left\|x_1\right\|_2^2 + \varepsilon \left\|x_2\right\|_2^2 \\
  \left\|x_1-x_2\right\|_2^2 &\leq \left(1+\frac{1}{\varepsilon}\right)\left\|x_1\right\|_2^2 + \left(1+\varepsilon\right)\left\|x_2\right\|_2^2
\end{aligned}
\end{equation}
\end{fact}

    

\subsection{Useful Lemmas}

\begin{lemma} \label{lm: lemma_drift} If $\tilde{\eta}^2 \le \frac{\beta^2R}{4(1+R)L_{\mathcal{L}}^2}$, the task drift error is upper-bounded 
\begin{equation}
\begin{aligned}
    & \mathbb{E} \Big[ \|g_{i,r}^t-\nabla \mathcal{L}_i(w^t)\|^2 \Big] \nonumber\\
    & \le 2 \lambda^2 \zeta^2 + \frac{32 \tilde{\eta}^2 L_{\mathcal{L}}^2}{\beta^2}
   \Big( \mathbb{E} \big[ \|\nabla \mathcal{L}_i(w^t)\|^2 \big] + 2\lambda^2 \zeta^2 \Big).
\end{aligned}
\end{equation}

\end{lemma}

\begin{lemma} \label{lm: lemma_deviation} If Assumption \ref{assumption_1_variance} holds and $\lambda > 2 \sqrt{2} L_{\ell}$, then we have
\begin{sloppypar}
   \begin{align}
        \frac{1}{N} \sum_{i=1}^{N} \big{\|}\nabla \mathcal{L}_i(w) - \nabla \mathcal{L}(w)\big{\|}^2 \le  2 \kappa_{\mathcal{L}}^2+ \frac{8L_{\ell}^2}{\lambda^2-8L_{\ell}^2}
         \big{\|}  \nabla \mathcal{L}(w) \big{\|}^2,
   \end{align}
\end{sloppypar}
\noindent where $\kappa_{\mathcal{L}}^2=\frac{{\lambda^2} \kappa_2^2}{{\lambda^2}-{8L_{\ell}^2}}$.
\end{lemma}

\subsection{Proof of Theorem \ref{thm: inexactsolution}}\label{proof_inexactsolution}

Since $\ell_{i}(\cdot)$ is $L_\ell$-smooth and $f_i(\cdot,w)$ is $\lambda$-strongly-convex, we get the $\rho_w^i(\cdot)$ is $(\lambda - L_\ell)$-strongly convex. Besides, we have that $\theta^\star_i(w)$ is the unique solution of $\rho_w^i(\theta)$ and $\nabla \rho_w^i(\theta^\star_i(w))=0$. Together with the strongly-convexity of $\rho_w^i(\theta)$, we obtain
\begin{align}
    & \norm{\hat{\theta}_i(w)-\theta^\star_i(w)}^2 \nonumber\\
    & \le \frac{1}{(\lambda - L_\ell)^2} \norm{\nabla \rho_w^i(\hat{\theta}_i(w))-\nabla \rho_w^i(\theta^\star_i(w))}^2 \nonumber \\
    & = \frac{1}{(\lambda - L_\ell)^2} \norm{\nabla \rho_w^i(\hat{\theta}_i(w))}^2 \nonumber\\
    &\le \!\frac{2}{(\!\lambda\! - \!L_\ell)^2} \!\Big(\!\norm{\nabla \rho_w^i(\hat{\theta}_i(w))\!-\!\nabla \tilde{\rho}_w^i(\hat{\theta}_i(w))}^2\! \!+\! \|\nabla \tilde{\rho}_w^i(\hat{\theta}_i(w))\|^2 \Big)\!\nonumber\\
    &\le \frac{2}{(\lambda - L_\ell)^2} \Big(\norm{\nabla \ell_i(\hat{\theta}_i(w))-\nabla \tilde{\ell}_i(\hat{\theta}_i(w))}^2 +  \delta^2 \Big),
\end{align}
where the second inequality uses Jensen's inequality and the last uses the assumption
$\|\nabla \tilde{\rho}_w^i(\hat{\theta}_i(w)) \| \le \delta$. Taking the expectation over the trajectories implies
\begin{align}
    \mathbb{E}\Big[\norm{\hat{\theta}_i(w)-\theta^\star_i(w)}^2\Big] \le \frac{2}{(\lambda - L_\ell)^2} \big(\kappa_1^2 +  \delta^2 \big),
\end{align}
which completes the proof.

\subsection{Proof of Theorem \ref{thm: convergence}}\label{proof_convergence}
\begin{proof}
    Since $\mathcal{L}(\cdot)$ is $L_{\mathcal{L}}$-smooth, then we have
    \begin{align}
     &\mathbb{E} [ \mathcal{L}(w^{t+1})-\mathcal{L}(w^t) ] \nonumber\\ 
        &\le \mathbb{E} [ \langle \nabla \mathcal{L}(w^t),w^{t+1}-w^t   \rangle] + \frac{L_{\mathcal{L}}}{2} \mathbb{E}\Big[\norm{w^{t+1}-w^t}^2\Big] \nonumber\\
        &= - \tilde{\eta} \mathbb{E} [ \langle \nabla \mathcal{L}(w^t), g^t \rangle] + \frac{ \tilde{\eta}^2 L_{\mathcal{L}}}{2} \mathbb{E}\Big[\norm{g^t}^2\Big] \nonumber\\
        &= - \tilde{\eta} \mathbb{E} [ \| \nabla \mathcal{L}(w^t)\|^2] - \tilde{\eta} \mathbb{E} [ \langle \nabla \mathcal{L}(w^t), g^t- \nabla \mathcal{L}(w^t)  \rangle] \nonumber\\
        & \quad + \frac{ \tilde{\eta}^2 L_{\mathcal{L}}}{2} \mathbb{E}\Big[\norm{g^t}^2\Big].
    \end{align}

By using the fact that $2|\ip{x}{y}|\le \|x\|^2+\|y\|^2$ for any vectors $x,y \in \mathbb{R}^d$ , we obtain
\begin{align} \label{neq:L2}
    & \mathbb{E} [ \mathcal{L}(w^{t+1})-\mathcal{L}(w^t) ] 
    \nonumber\\
    & \le -\frac{\tilde{\eta}}{2} \mathbb{E} [ \| \nabla \mathcal{L}(w^t)\|^2] 
    + \frac{\tilde{\eta}}{2} \mathbb{E} [ \|g^t- \nabla \mathcal{L}(w^t) \|^2] \nonumber\\
    & \quad + \frac{ \tilde{\eta}^2 L_{\mathcal{L}}}{2} \mathbb{E}\Big[\norm{g^t}^2\Big].
\end{align}

By using AM-GM inequality and Jensen's inequality, we have
\begin{align} \label{neq:gt}
    \norm{g^t}^2 \le & 3 \bigg[\bigg{\|}\frac{1}{NR} \sum_{i,r=1}^{N,R} g_{i,r}^t-\frac{1}{N} \sum_{i=1}^{N}\nabla \mathcal{L}_i(w^t) \bigg{\|}^2 \nonumber\\ 
    & + \bigg{\|}\frac{1}{N} \sum_{i=1}^{N}\nabla \mathcal{L}_i(w^t) - \nabla \mathcal{L}(w^t)\bigg{\|}^2 + \bigg{\|}\nabla \mathcal{L}(w^t)\bigg{\|}^2 \bigg] \nonumber\\
    \le& \frac{3}{NR} \sum_{i,r=1}^{N,R} \|g_{i,r}^t-\nabla \mathcal{L}_i(w^t)\|^2 \nonumber\\
    & + \frac{3}{N} \sum_{i=1}^{N} \bigg{\|}\nabla \mathcal{L}_i(w^t) - \nabla \mathcal{L}(w^t)\bigg{\|}^2
    + 3 \bigg{\|}\nabla \mathcal{L}(w^t)\bigg{\|}^2. 
\end{align}

Incorporating \eqref{neq:gt} into \eqref{neq:L2} and using Jensen's inequality, we get
\begin{align} \label{neq:L3}
     &\mathbb{E} [ \mathcal{L}(w^{t+1})-\mathcal{L}(w^t) ]  
     \nonumber\\
    &\le-\frac{\tilde{\eta} (1-3L_{\mathcal{L}}\tilde{\eta}) }{2} \mathbb{E} [ \| \nabla \mathcal{L}(w^t)\|^2]  \nonumber\\
    & \quad + \frac{3 \tilde{\eta}^2 L_{\mathcal{L}}}{2} \frac{1}{N} \sum_{i=1}^{N} \mathbb{E} \Big[ \big{\|}\nabla \mathcal{L}_i(w^t) - \nabla \mathcal{L}(w^t)\big{\|}^2 \Big] \nonumber\\
    & \quad +\frac{\tilde{\eta}(1+3\tilde{\eta}L_{\mathcal{L}})}{2} \frac{1}{NR} \sum_{i,r=1}^{N,R} \mathbb{E} \Big[ \|g_{i,r}^t-\nabla \mathcal{L}_i(w^t)\|^2 \Big].
\end{align}

By using Lemma \ref{lm: lemma_drift}, we obtain 
\begin{align} \label{neq:L4}
     & \mathbb{E} [ \mathcal{L}(w^{t+1})-\mathcal{L}(w^t) ]  
     \nonumber\\
    & \le -\frac{\tilde{\eta} (1-3L_{\mathcal{L}}\tilde{\eta}) }{2} \mathbb{E} [ \| \nabla \mathcal{L}(w^t)\|^2] \nonumber\\
    & \quad + \frac{3 \tilde{\eta}^2 L_{\mathcal{L}}}{2} \frac{1}{N} \sum_{i=1}^{N} \mathbb{E} \Big[ \big{\|}\nabla \mathcal{L}_i(w^t) - \nabla \mathcal{L}(w^t)\big{\|}^2 \Big] \nonumber\\
    & \quad +\frac{\tilde{\eta}(1+3\tilde{\eta}L_{\mathcal{L}})}{2} \bigg(2 \lambda^2 \zeta^2 \nonumber\\
    & \quad +\frac{32 \tilde{\eta}^2 L_{\mathcal{L}}^2}{\beta^2}
   \Big( \frac{1}{N} \sum_{i=1}^{N}  \mathbb{E} \big[ \|\nabla \mathcal{L}_i(w^t)\|^2 \big] + 2\lambda^2 \zeta^2 \Big)\bigg). 
\end{align}

Using the fact that $\frac{1}{N}\sum_{i=1}^N \norm{x_i}^2 = \norm{\bar{x}}^2+\frac{1}{N}\sum_{i=1}^N \norm{x_i-\bar{x}}^2$, where $\bar{x}=\frac{1}{N}\sum_{i=1}^N x_i$, we get 

\begin{align}
 \label{neq:L5}
     &\mathbb{E} [ \mathcal{L}(w^{t+1})-\mathcal{L}(w^t) ] \nonumber\\
     & \le -\frac{\tilde{\eta} (1-3L_{\mathcal{L}}\tilde{\eta}) }{2} \mathbb{E} [ \| \nabla \mathcal{L}(w^t)\|^2] \nonumber\\ 
    & \quad + \frac{3 \tilde{\eta}^2 L_{\mathcal{L}}}{2} \frac{1}{N} \sum_{i=1}^{N} \mathbb{E} \Big[ \big{\|}\nabla \mathcal{L}_i(w^t) - \nabla \mathcal{L}(w^t)\big{\|}^2 \Big] 
    \nonumber\\
    & \quad +\frac{\tilde{\eta}(1+3\tilde{\eta}L_{\mathcal{L}})}{2} \bigg(2 \lambda^2 \zeta^2 + \frac{32 \tilde{\eta}^2 L_{\mathcal{L}}^2}{\beta^2}
   \Big( \frac{1}{N} \sum_{i=1}^{N}  \mathbb{E} \Big[ \|\nabla \mathcal{L}_i(w^t) \nonumber\\
   & \quad - \nabla \mathcal{L}(w^t)\|^2 \Big] + \mathbb{E} \big[ \|\nabla \mathcal{L}(w^t)\|^2 \big] + 2\lambda^2 \zeta^2 \Big)\bigg)\nonumber\\
   & = \tilde{\eta}(1+3\tilde{\eta}L_{\mathcal{L}})\lambda^2 \zeta^2 \bigg( 1 + \frac{32 \tilde{\eta}^2 L_{\mathcal{L}}^2}{\beta^2}  \bigg)\nonumber\\
   & \quad + \bigg( \frac{16 \tilde{\eta}^3 (1+3\tilde{\eta}L_{\mathcal{L}}) L_{\mathcal{L}}^2}{\beta^2} -\frac{\tilde{\eta} (1-3L_{\mathcal{L}}\tilde{\eta}) }{2} \bigg) \mathbb{E} [ \| \nabla \mathcal{L}(w^t)\|^2] \nonumber\\
   & \quad + \bigg( \frac{16 \tilde{\eta}^3 (1+3\tilde{\eta}L_{\mathcal{L}}) L_{\mathcal{L}}^2}{\beta^2} + \frac{3 \tilde{\eta}^2 L_{\mathcal{L}}}{2}  \bigg) \bigg(\frac{1}{N} \sum_{i=1}^{N} \mathbb{E} \Big[ \big{\|}\nabla \mathcal{L}_i(w^t) \nonumber\\
   & \quad - \nabla \mathcal{L}(w^t)\big{\|}^2 \Big]\bigg). \nonumber\\
\end{align}

By using Lemma \ref{lm: lemma_deviation}, we get 
\begin{align}
 \label{neq:L6}
     & \mathbb{E} [ \mathcal{L}(w^{t+1})-\mathcal{L}(w^t) ] \nonumber\\ 
    & \le  \tilde{\eta}(1+3\tilde{\eta}L_{\mathcal{L}})\lambda^2 \zeta^2 \bigg( 1 + \frac{32 \tilde{\eta}^2 L_{\mathcal{L}}^2}{\beta^2}  \bigg)
    \nonumber\\
   & \quad + \bigg( \frac{32 \tilde{\eta}^3 (1+3\tilde{\eta}L_{\mathcal{L}}) L_{\mathcal{L}}^2}{\beta^2} +  3\tilde{\eta}^2 L_{\mathcal{L}} \bigg)  \kappa_{\mathcal{L}}^2 \nonumber\\
   & \quad + \tilde{\eta} \bigg[ \frac{16 \tilde{\eta}^2 (1+3\tilde{\eta}L_{\mathcal{L}}) L_{\mathcal{L}}^2}{\beta^2} \frac{\lambda^2}{\lambda^2-8L_{\ell}^2} \nonumber\\
   & \quad - \frac{ (1-3L_{\mathcal{L}}\tilde{\eta}) }{2} + \frac{12\tilde{\eta} L_{\mathcal{L}} L_{\ell}^2}{\lambda^2-8L_{\ell}^2} \bigg] \mathbb{E} [ \| \nabla \mathcal{L}(w^t)\|^2]  \nonumber\\
   & \le  2 \tilde{\eta}\lambda^2 \zeta^2 + 3 \tilde{\eta}^2 L_{\mathcal{L}}   \kappa_{\mathcal{L}}^2 + \frac{64\tilde{\eta}^3 L_{\mathcal{L}}^2 (\lambda^2 \zeta^2 + \kappa_{\mathcal{L}}^2) }{\beta^2} \nonumber\\
   & \quad - \tilde{\eta} \bigg[ \! \frac{1}{2} \! - \!\tilde{\eta} L_{\mathcal{L}} \bigg(\!\frac{3}{2}\!+\! \frac{16\lambda^2}{\lambda^2-8L_{\ell}^2} \! + \!  \frac{12 L_{\ell}^2}{\lambda^2-8L_{\ell}^2} \! \bigg) \! \bigg] \mathbb{E} [ \| \nabla \mathcal{L}(w^t)\|^2],\nonumber\\
\end{align}
where the second inequality uses $\tilde{\eta}\le \frac{\beta}{2L_{\mathcal{L}}}$, $\beta \ge 1$ and $1+3\tilde{\eta}L_{\mathcal{L}}\le 1+\frac{3}{70 \lambda^2} < 2$ since $\tilde{\eta} \le \frac{1}{70L_{\mathcal{L}} \lambda^2}$ and $\lambda^2-8L_{\ell}^2\ge 1$.
Using $\lambda^2-8L_{\ell}^2\ge 1$ again implies
\begin{align}
    &L_{\mathcal{L}} \bigg(\frac{3}{2}+ \frac{16\lambda^2}{\lambda^2-8L_{\ell}^2} + \frac{12 L_{\ell}^2}{\lambda^2-8L_{\ell}^2} \bigg) \nonumber\\
   &\le  L_{\mathcal{L}} \bigg(\frac{3}{2}+ 16\lambda^2 +  12 L_{\ell}^2 \bigg) \nonumber\\
   &\le \frac{35}{2} L_{\mathcal{L}} \lambda^2.
\end{align}
Using the assumption that $\tilde{\eta} \le \frac{1}{70L_{\mathcal{L}} \lambda^2}\triangleq \tilde{\eta}_o$ yields
\begin{align}
     \frac{1}{2} -  \tilde{\eta} L_{\mathcal{L}} \bigg(\frac{3}{2}+ \frac{16\lambda^2}{\lambda^2-8L_{\ell}^2} +  \frac{12 L_{\ell}^2}{\lambda^2-8L_{\ell}^2} \bigg) \ge \frac{1}{4},
\end{align}
then we obtain
\begin{align} \label{neq:L7}
     & \mathbb{E} [ \mathcal{L}(w^{t+1})-\mathcal{L}(w^t)   
     \nonumber\\
   & \le  2 \tilde{\eta}\lambda^2 \zeta^2 + 3 \tilde{\eta}^2 L_{\mathcal{L}}   \kappa_{\mathcal{L}}^2  + \frac{64\tilde{\eta}^3 L_{\mathcal{L}}^2 (\lambda^2 \zeta^2 + \kappa_{\mathcal{L}}^2) }{\beta^2} \nonumber\\
   & \quad - \frac{\tilde{\eta}}{4}  \mathbb{E} [ \| \nabla \mathcal{L}(w^t)\|^2],
\end{align}

By reorganizing the above inequality and telescoping, we obtain

\begin{align} \label{neq:L8}
     & \frac{1}{4T} \sum_{t=1}^{T} \mathbb{E} [ \| \nabla \mathcal{L}(w^t)\|^2] \nonumber\\
     & \le  \frac{\mathbb{E} [ \mathcal{L}(w^{1})-\mathcal{L}(w^{T+1}) ]} {\tilde{\eta}T} \nonumber\\
   & \quad +  \frac{64\tilde{\eta}^2 L_{\mathcal{L}}^2 (\lambda^2 \zeta^2 + \kappa_{\mathcal{L}}^2) }{\beta^2} + 3 \tilde{\eta} L_{\mathcal{L}}   \kappa_{\mathcal{L}}^2 + 2\lambda^2 \zeta^2 ,
\end{align}

Let $w^\star$ be the optimal solution of $\mathcal{L}(w)$ and define $\Delta \triangleq \mathcal{L}(w^{1})-\mathcal{L}(w^\star)$. It's obvious that 
\begin{equation}
     \Delta \triangleq \mathcal{L}(w^{1})-\mathcal{L}(w^\star) \ge \mathbb{E} [ \mathcal{L}(w^{1})-\mathcal{L}(w^{T+1}) ].
\end{equation}

We discuss the upper bound in two cases: 1) if $\tilde{\eta}_o \ge \min \Big\{\sqrt{\frac{\Delta}{3T L_{\mathcal{L}} \kappa_{\mathcal{L}}^2}}, \sqrt[3]{\frac{\beta^2\Delta}{64T L_{\mathcal{L}}^2 (\lambda^2 \zeta^2 + \kappa_{\mathcal{L}}^2) }} \Big\} \triangleq c_o $, choosing $\tilde{\eta}= c_o$ implies

\begin{align} \label{neq:L9}
   & \frac{1}{4T} \sum_{t=1}^{T} \mathbb{E} [ \| \nabla \mathcal{L}(w^t)\|^2] \nonumber\\
   & \le 2 \sqrt[3]{\frac{64\Delta^2 L_{\mathcal{L}}^2 (\lambda^2 \zeta^2 
    + \kappa_{\mathcal{L}}^2) }{\beta^2 T^2}} + 2 \sqrt{\frac{3 \Delta L_{\mathcal{L}}   \kappa_{\mathcal{L}}^2}{T}} + 2\lambda^2 \zeta^2
\end{align}

2) if $\tilde{\eta}_o\le c_o$, choosing $\tilde{\eta}=\tilde{\eta}_o$ implies
\begin{align} \label{neq:L10}
     & \frac{1}{4T} \sum_{t=1}^{T} \mathbb{E} [ \| \nabla \mathcal{L}(w^t)\|^2] \nonumber\\
     & \le \frac{\Delta}{\tilde{\eta}_o T} + \sqrt[3]{\frac{64\Delta^2 L_{\mathcal{L}}^2 (\lambda^2 \zeta^2 + \kappa_{\mathcal{L}}^2) }{\beta^2 T^2}} \nonumber\\
     & \quad + \sqrt{\frac{3 \Delta L_{\mathcal{L}}   \kappa_{\mathcal{L}}^2}{T}} + 2\lambda^2 \zeta^2.
\end{align}
\end{proof}

Therefore, combining \eqref{neq:L9} and \eqref{neq:L10} implies that
\begin{align}
    & \frac{1}{T} \sum_{t=1}^{T} \mathbb{E} [ \| \nabla \mathcal{L}(w^t)\|^2] = \mathbb{E} [ \| \nabla \mathcal{L}(w^{\tilde{t}})\|^2] \nonumber\\
    & \le \mathcal{O} \bigg(\frac{\Delta}{\tilde{\eta}_o T} + \sqrt[3]{\frac{\Delta^2 L_{\mathcal{L}}^2 (\lambda^2 \zeta^2 + \kappa_{\mathcal{L}}^2) }{\beta^2 T^2}} + \sqrt{\frac{\Delta L_{\mathcal{L}}   \kappa_{\mathcal{L}}^2}{T}} + \lambda^2 \zeta^2\bigg),
\end{align}
where $\tilde{t}$ is sampled uniformly from the index set $\{1,\ldots,T\}$.

Next, we obtain
\begin{align} \label{neq:L11_1}
    & \frac{1}{N} \sum_{i=1}^N \mathbb{E}\Big[\|\hat{\theta}_i(w_t)-w_t\|^2 \Big] \nonumber\\
    &\le \! \frac{1}{N} \! \sum_{i=1}^N 2 \!\bigg(\!\mathbb{E}\Big[\|\hat{\theta}_i(w_t)\! - \! \theta^\star(w_t)\|^2 \Big]\!+\!\mathbb{E}\Big[\|w_t-\theta^\star(w_t)\|^2 \Big] \! \bigg) \nonumber \\
    & \le 2 \zeta^2 + \frac{2}{N} \sum_{i=1}^N \frac{\mathbb{E}\Big[\|\nabla \mathcal{L}_i(w_t)\|^2 \Big]}{\lambda^2}\nonumber\\
    & = 2 \zeta^2 + \frac{2}{N} \sum_{i=1}^N \frac{\mathbb{E}\Big[\|\nabla \mathcal{L}_i(w_t)-\nabla \mathcal{L}(w_t)\|^2 + \|\nabla \mathcal{L}(w_t)\|^2\Big]}{\lambda^2}\nonumber \\
    & \le 2 \zeta^2 + \frac{2 \kappa_{\mathcal{L}}^2}{\lambda^2}+\frac{2}{\lambda^2-8L_{\ell}^2} \mathbb{E} [ \| \nabla \mathcal{L}(w^t)\|^2],
\end{align}
where the first inequality applies Jensen's inequality, the second inequality uses the $\zeta$-approximation and Eq. \eqref{eq: Gradient_L}, the third line utilizes the fact that $\frac{1}{N}\sum_{i=1}^N \norm{x_i}^2 = \norm{\bar{x}}^2+\frac{1}{N}\sum_{i=1}^N \norm{x_i-\bar{x}}^2$, and the last inequality follows from Lemma \ref{lm: lemma_deviation}.

By taking the average over $t$ from $1$ to $T$ for Eq. \eqref{neq:L11_1}, we get
\begin{align} \label{neq:L11_2}
    & \frac{1}{TN} \sum_{t=1}^T \sum_{i=1}^N \mathbb{E}\Big[\|\hat{\theta}_i(w_t)-w_t\|^2 \Big] \nonumber\\
    & \le  2 \zeta^2 + \frac{2 \kappa_{\mathcal{L}}^2}{\lambda^2}+\frac{2}{\lambda^2-8L_{\ell}^2} \frac{1}{T} \sum_{t=1}^T  \mathbb{E} [ \| \nabla \mathcal{L}(w^t)\|^2].
\end{align}

Together with Eq. \eqref{neq:L10}, we finish the proof
\begin{align} \label{neq:L12}
    & \frac{1}{N} \sum_{i=1}^N \mathbb{E}\Big[\|\hat{\theta}_i(w_{\tilde{t}})-w_{\tilde{t}}\|^2 \Big] \nonumber\\
    & \le \mathcal{O} \bigg(\zeta^2 + \frac{ \kappa_{\mathcal{L}}^2}{\lambda^2} \bigg) + \mathcal{O} \Big({\mathbb{E} \big[\| \nabla \mathcal{L}(w^{\tilde{t}})\|^2 \big]} \Big).
\end{align}

\subsection{Proof of Important Lemmas}
\subsubsection{Proof of Lemma \ref{lm: smooth_L_l}}
    Define 
    \begin{align}
    & \delta_F(\theta_i,s') \nonumber\\
    & = \sum_{s_i, a_i: T_i(s_i, a_i)= s_i'} F_{\theta_i}(s_i, a_i) - R (s_i') - \sum_{a_i' \in \mathcal{A}(s_i')} F_{\theta_i}\left(s_i', a_i'\right), \nonumber\\
    & \delta_{\nabla F}(\theta_i,s') \nonumber\\
    & = \sum_{s_i, a_i: T_i(s_i, a_i)= s_i'} \nabla_{\theta_i} F_{\theta_i}(s_i, a_i)  - \sum_{a_i' \in \mathcal{A}(s_i')} \nabla_{\theta_i}  F_{\theta_i}\left(s_i', a_i'\right).
    \end{align}
      
    For each $\theta_i,\theta_i^\prime \in \mathbb{R}^d$, using triangle inequality yields
    \begin{align} \label{smooth-1}
       & \|\nabla \ell_{i}(\theta_i)-\nabla \ell_{i}(\theta'_i)\| \nonumber\\
        & = \mathbb{E}_{\tau_i \sim p_i(\tau)} \Big[ \sum_{s_i' \in \tau_i \neq s_{0}} \Big(\delta_F(\theta_i,s')\delta_{\nabla F}(\theta_i,s') \nonumber\\
        & \quad -\delta_F(\theta_i^\prime,s')\delta_{\nabla F}(\theta_i^\prime,s')\Big)\Big] \nonumber\\
        & = \mathbb{E}_{\tau_i \sim p_i(\tau)} \Big[ \sum_{s_i' \in \tau_i \neq s_{0}} \Big(\delta_F(\theta_i,s')\delta_{\nabla F}(\theta_i,s') \nonumber\\
        & \quad -\delta_F(\theta_i,s')\delta_{\nabla F}(\theta_i',s') +\delta_F(\theta_i,s')\delta_{\nabla F}(\theta_i',s') \nonumber \\
        & \quad -\delta_F(\theta_i^\prime,s')\delta_{\nabla F}(\theta_i^\prime,s')\Big)\Big] \nonumber\\
        & \le \mathbb{E}_{\tau_i \sim p_i(\tau)} \Big[ \sum_{s_i' \in \tau_i \neq s_{0}} \Big( \big{\|}\delta_F(\theta_i,s') \cdot \big[\delta_{\nabla F}(\theta_i,s') \nonumber\\
        & \quad -\delta_{\nabla F}(\theta_i',s')\big] \big{\|}  + \big{\|} \big[\delta_F(\theta_i,s')\nonumber\\
        & \quad -\delta_F(\theta_i^\prime,s') \big]\delta_{\nabla F}(\theta_i^\prime,s')\big{\|} \Big)\Big].
    \end{align}
Under Assumptions \ref{assum:F_theta}, \ref{assum: limited state} and \ref{assum: reward}, we have 
\begin{align}
\|\delta_F(\theta_i,s')\| &\le H_0+2H_2 B, \label{smooth-n-1}\\
\|\delta_{\nabla F}(\theta_i,s')-\delta_{\nabla F}(\theta_i',s')\| &\le 2 H_2 L_1 \|\theta_i-\theta_i^\prime\|, \label{smooth-n-2}\\
\|\delta_F(\theta_i,s')-\delta_F(\theta_i^\prime,s')\| & \le 2H_2 L_0 \|\theta_i-\theta_i^\prime\|,\label{smooth-n-3}\\
\|\delta_{\nabla F}(\theta_i^\prime,s')\| &\le 2H_2 L_0. \label{smooth-n-4}
\end{align}

Combining (\ref{smooth-n-1}-\ref{smooth-n-4}) with \eqref{smooth-1}, we obtain
\begin{align}
 \label{smooth-2}
        & \|\nabla \ell_{i}(\theta_i)-\nabla \ell_{i}(\theta'_i)\| \nonumber\\
        & \le H_1\Big[(H_0+2H_2 B)2 H_2 L_1 \|\theta_i-\theta_i^\prime\| \nonumber\\
        & \quad + 2H_2 L_0 \|\theta_i-\theta_i^\prime\| 2H_2 L_0 \Big] \nonumber \\
        & = H_1\Big[(H_0+2H_2 B)2 H_2 L_1+ 4H_2^2 L_0^2\Big] \|\theta_i-\theta_i^\prime\|.
\end{align} 

\subsubsection{Proof of Lemma \ref{lm: smooth_L_L}}
    See the proof in Corollary 3.4 b) of \cite{hoheisel2020regularization}.

\subsubsection{Proof of lemma \ref{lm: lemma_drift}}

By using Jensen’s inequality, the smoothness of $\nabla \mathcal{L}_i$ and $\zeta$-approximation between $\hat{\theta}_i(w^t_{i,r})$ and $\theta_i^\star(w^t_{i,r})$, we obtain
\begin{align} \label{eq: task_drift_1}
     & \mathbb{E} \Big[ \|g_{i,r}^t-\nabla \mathcal{L}_i(w^t)\|^2 \Big] \nonumber\\ 
     & \le 2  \mathbb{E} \Big[ \|g_{i,r}^t-\nabla \mathcal{L}_i(w^t_{i,r})\|^2 + \|\nabla \mathcal{L}_i(w^t_{i,r})-\nabla \mathcal{L}_i(w^t)\|^2 \Big] \nonumber\\
     & \le 2 \lambda^2 \mathbb{E}\big[ \|\hat{\theta}_i(w^t_{i,r}) - \theta_i^\star(w^t_{i,r})\|^2\big] +2 L_{\mathcal{L}}^2 \mathbb{E} \big[ \| w^t_{i,r}-w^t\|^2\big] \nonumber \\
     & \le  2 \lambda^2 \zeta^2 +2 L_{\mathcal{L}}^2 \mathbb{E} \big[ \| w^t_{i,r}-w^t\|^2\big].
\end{align}
Next, we give the upper bound of $\mathbb{E} \big[ \| w^t_{i,r}-w^t\|^2\big]$. Replacing $w^t_{i,r}$ with its update rule \eqref{eq: wt_ir_update}, using Young's inequality and Jensen's inequality, for any $q$, we have
\begin{align} \label{eq: task_drift_2}
    &\mathbb{E} \big[ \| w^t_{i,r}-w^t\|^2\big]\nonumber\\
    &= \mathbb{E} \big[ \| w^t_{i,r-1}- \eta  g_{i,r-1}^t - w^t\|^2\big] \nonumber \\
    &\le \left(1+\frac{1}{q} \right) \mathbb{E} \big[ \| w^t_{i,r-1} - w^t\|^2\big] 
    +(1+q) \eta^2 \mathbb{E} \big[ \| g_{i,r-1}^t\|^2\big] \nonumber \\
    &\le \left(1+\frac{1}{q} \right) \mathbb{E} \big[ \| w^t_{i,r-1} - w^t\|^2\big] \nonumber\\
    & \quad +2 (1+q) \eta^2  \mathbb{E} \big[ \|\nabla \mathcal{L}_i(w^t)\|^2 \big] \nonumber\\
    & \quad + 2 (1+q) \eta^2 \mathbb{E} \big[ \| g_{i,r-1}^t- \nabla \mathcal{L}_i(w^t) \|^2\big]. 
\end{align}

Incorporating Eq. \eqref{eq: task_drift_1} into \eqref{eq: task_drift_2}, we obtain
\begin{align} \label{eq: task_drift_3}
    &\mathbb{E} \big[ \| w^t_{i,r}-w^t\|^2\big] \nonumber\\
    &\le \left(1+\frac{1}{q} \right) \mathbb{E} \big[ \| w^t_{i,r-1} - w^t\|^2\big] 
    +2 (1+q) \eta^2  \mathbb{E} \big[ \|\nabla \mathcal{L}_i(w^t)\|^2 \big]\nonumber\\
    & \quad + 4 (1+q) \eta^2 (\lambda^2 \zeta^2 +L_{\mathcal{L}}^2 \mathbb{E} \big[ \| w^t_{i,r-1}-w^t\|^2\big]) \nonumber\\
    & = \left(1+\frac{1}{q} +4 (1+q) \eta^2 L_{\mathcal{L}}^2 \right) \mathbb{E} \big[ \| w^t_{i,r-1} - w^t\|^2\big] \nonumber\\
    & \quad +2 (1+q) \eta^2 \mathbb{E} \big[ \|\nabla \mathcal{L}_i(w^t)\|^2 \big] +4 (1+q) \eta^2 \lambda^2 \zeta^2.
\end{align}

Let $q=R$. Under the condition that $\tilde{\eta}^2 \le \frac{\beta^2R}{4(1+R)L_{\mathcal{L}}^2}$, we have $4 (1+R) \eta^2 L_{\mathcal{L}}^2= 4 (1+R) L_{\mathcal{L}}^2 \frac{\tilde{\eta}^2}{\beta^2 R^2} \le \frac{1}{R}$. Then we get
\begin{align} \label{eq: task_drift_4}
    &\mathbb{E} \big[ \| w^t_{i,r}-w^t\|^2\big] \nonumber\\
    &\le
    \left(1+\frac{2}{R} \right) \mathbb{E} \big[ \| w^t_{i,r-1} - w^t\|^2\big] \nonumber\\
    & \quad +\frac{4 \tilde{\eta}^2}{\beta^2 R} \mathbb{E} \big[ \|\nabla \mathcal{L}_i(w^t)\|^2 \big] + \frac{8\tilde{\eta}^2\lambda^2 \zeta^2 }{\beta^2 R} \nonumber\\
   & \le \frac{4 \tilde{\eta}^2}{\beta^2 R}
   \bigg( \mathbb{E} \big[ \|\nabla \mathcal{L}_i(w^t)\|^2 \big] + 2\lambda^2 \zeta^2 \bigg) \sum_{r=0}^{R-1}  \left(1+\frac{2}{R} \right)^r\nonumber\\
   &\le \frac{16 \tilde{\eta}^2}{\beta^2}
   \bigg( \mathbb{E} \big[ \|\nabla \mathcal{L}_i(w^t)\|^2 \big] + 2\lambda^2 \zeta^2 \bigg),
\end{align}
where the last inequality applies that $(1+a/n)^n\le e^a$ for all $a \in \mathbb{R}$ and $n \in \mathbb{N}$ and $\sum_{r=0}^{R-1}  \left(1+\frac{2}{R} \right)^r=\frac{(1+2/R)^R-1}{2/R}\le \frac{e^2-1}{2/R} < 4 R$. Combining Eqs. \eqref{eq: task_drift_1} and \eqref{eq: task_drift_4} completes the proof.

\subsubsection{Proof of lemma \ref{lm: lemma_deviation}}
   Since the gradient of $\mathcal{L}_i(w)$ is  $ \nabla \mathcal{L}_i(w) = \lambda (w-\theta_i^\star(w))$, we have 
   \begin{align}
         & \big{\|}\nabla \mathcal{L}_i(w) - \nabla \mathcal{L}(w)\big{\|}^2  \nonumber\\
         & =  \bigg{\|} \lambda (w-\theta_i^\star(w)) - \frac{1}{N} \sum_{k=1}^N \lambda (w-\theta_k^\star(w)) \bigg{\|}^2 
   \end{align}
   The first-order optimal condition of Eq. \eqref{pMeta-GFN-a} implies that 
   $\nabla \ell_i( \theta_i^\star (w) ) -\lambda (w-\theta_i^\star (w)) =0$. Together with Jensen's inequality, we have
   \begin{align}
         &\big{\|}\nabla \mathcal{L}_i(w) - \nabla \mathcal{L}(w)\big{\|}^2 \nonumber\\ 
         &=  \bigg{\|} \nabla \ell_i( \theta_i^\star (w) ) - \frac{1}{N} \sum_{j=1}^N \nabla \ell_j( \theta_j^\star (w) ) \bigg{\|}^2 \\
         &\le 2  \bigg{\|} \nabla \ell_i( \theta_i^\star (w) ) - \frac{1}{N} \sum_{j=1}^N \nabla \ell_j( \theta_i^\star (w) ) \bigg{\|}^2 \nonumber\\
         & \quad +  2 \bigg{\|} \frac{1}{N} \sum_{j=1}^N \nabla \ell_j( \theta_i^\star (w) )  - \frac{1}{N} \sum_{j=1}^N \nabla \ell_j( \theta_j^\star (w) ) \bigg{\|}^2.
   \end{align}
   By taking the average over $N$ tasks, we obtain
   \begin{align} 
        &\frac{1}{N} \sum_{i=1}^{N} \big{\|}\nabla \mathcal{L}_i(w) - \nabla \mathcal{L}(w)\big{\|}^2 \nonumber\\ 
        &\le 2 \kappa_2^2 +
        \frac{2}{N^2} \sum_{i=1}^N \sum_{j=1}^N \bigg{\|}  \nabla \ell_j( \theta_i^\star (w) )  -  \nabla \ell_j( \theta_j^\star (w) ) \bigg{\|}^2 \nonumber\\
        & \le 2 \kappa_2^2 +
        \frac{2L_{\ell}^2}{N^2} \sum_{i=1}^N \sum_{j=1}^N \big{\|}  \theta_i^\star (w)  - \theta_j^\star (w)  \big{\|}^2 \nonumber\\
        &\le 2 \kappa_2^2 +
        \frac{2L_{\ell}^2}{N^2} \sum_{i=1}^N \sum_{j=1}^N \Big[\!2\big{\|}  \theta_i^\star (w)  \!-\! w  \big{\|}^2 +2 \big{\|}  \theta_i^\star (w) \! -\! w  \big{\|}^2\!\Big] \nonumber\\
        &= 2 \kappa_2^2 +
        \frac{4L_{\ell}^2}{N^2\lambda^2} \sum_{i=1}^N \sum_{j=1}^N \Big[  \big{\|} \nabla \mathcal{L}_i(w) \big{\|}^2 +\big{\|}  \nabla \mathcal{L}_j(w)  \big{\|}^2\Big] \nonumber\\
        &=2 \kappa_2^2 +
        \frac{8L_{\ell}^2}{N\lambda^2} \sum_{i=1}^N  \big{\|}  \nabla \mathcal{L}_i(w) \big{\|}^2 \nonumber\\
        &=2 \kappa_2^2 \!+\!
        \frac{8L_{\ell}^2}{\lambda^2} \bigg[\! \frac{1}{N}\sum_{i=1}^N  \big{\|}  \nabla \mathcal{L}_i(w) \!-\!\nabla \mathcal{L}(w) \big{\|}^2 \!+ \!\big{\|}  \nabla \mathcal{L}(w) \big{\|}^2\!\bigg]
   \end{align}
   where the first inequality uses the assumption \eqref{assumption_1_variance_sg} and Jensen's inequality, the second inequality follows from Lemma \ref{lm: smooth_L_l}, the third inequality applies Jensen's inequality, the first equality uses  $ \nabla \mathcal{L}_i(w) = \lambda (w-\theta_i^\star(w))$ and the second equality uses the fact that $\frac{1}{N}\sum_{i=1}^N \norm{x_i}^2 = \norm{\bar{x}}^2+\frac{1}{N}\sum_{i=1}^N \norm{x_i-\bar{x}}^2$, where $\bar{x}=\frac{1}{N}\sum_{i=1}^N x_i$.  When $\lambda > 2 \sqrt{2} L_{\ell}$, reorganizing the above inequality yields
   \begin{align}
        & \frac{1}{N} \sum_{i=1}^{N} \big{\|}\nabla \mathcal{L}_i(w) - \nabla \mathcal{L}(w)\big{\|}^2 \nonumber\\
        & \le \frac{2 {\lambda^2} \kappa_2^2}{{\lambda^2}-{8L_{\ell}^2}} + \frac{8L_{\ell}^2}{\lambda^2-8L_{\ell}^2}
         \big{\|}  \nabla \mathcal{L}(w) \big{\|}^2.
   \end{align}
}

%
\bibliographystyle{cas-model2-names}
\bibliography{example_paper.bib}

\end{document}